\documentclass[9pt,journal,compsoc]{IEEEtran}

\newcommand\tac[1]{\textcolor{black}{#1}}

\usepackage{times}
\usepackage{epsfig}
\usepackage{graphicx}
\usepackage{amsmath}
\usepackage{amssymb}
\usepackage{verbatim}
\usepackage{times}
\usepackage{svg}
\usepackage{epsfig}
\usepackage{adjustbox}
\usepackage{colortbl}
\usepackage{amsfonts}

\usepackage{pifont}

\usepackage{tikz}
\usepackage{pgfplots}
\pgfplotsset{compat=1.15}
\usepackage{amssymb}
\usetikzlibrary{positioning, calc, arrows, arrows.meta}
\usepgfplotslibrary{groupplots}

\tikzstyle{taskfeature}=[line width=0.8pt]
\tikzstyle{taskfeature1}=[taskfeature, fill=black!15!blue, fill opacity=0.4, color=black!30!blue]
\tikzstyle{taskfeature2}=[taskfeature, fill=black!20!green, fill opacity=0.4, color=black!30!green]
\tikzstyle{taskfeature3}=[taskfeature, fill=black!15!red, fill opacity=0.4, color=black!30!red]

\tikzstyle{task}=[line width=1pt, rounded corners, text opacity=1.0]
\tikzstyle{task1}=[fill=black!15!blue, task, fill opacity=0.4, color=black!30!blue]
\tikzstyle{task2}=[fill=black!20!green, task, fill opacity=0.4, color=black!30!green]
\tikzstyle{task3}=[fill=black!15!red, task, fill opacity=0.4, color=black!30!red]
\tikzstyle{positive}=[fill=black!15!green, fill opacity=0.5]
\tikzstyle{negative}=[fill=black!15!red, fill opacity=0.5]
\tikzstyle{neutral}=[fill=lightgray]

\tikzstyle{featbox}=[rectangle, minimum width=1em, minimum height=1em]
\tikzset{%
  do path picture/.style={%
    path picture={%
      \pgfpointdiff{\pgfpointanchor{path picture bounding box}{south west}}%
        {\pgfpointanchor{path picture bounding box}{north east}}%
      \pgfgetlastxy\x\y%
      \tikzset{x=\x/2,y=\y/2}%
      #1
    }
  },
  sin wave/.style={do path picture={    
    \draw [line cap=round] (-3/4,0)
      sin (-3/8,1/2) cos (0,0) sin (3/8,-1/2) cos (3/4,0);
  }},
  cross/.style={do path picture={    
    \draw [line cap=round] (-1,-1) -- (1,1) (-1,1) -- (1,-1);
  }},
  plus/.style={do path picture={    
    \draw [line cap=round] (-3/4,0) -- (3/4,0) (0,-3/4) -- (0,3/4);
  }}
}

\def\vertspace{1}
\def\horspace{3}

\usepackage{algorithmic}
\usepackage{algorithm}
\ifCLASSOPTIONcompsoc
  \usepackage[nocompress, noadjust]{cite}
\else
  \usepackage{cite}
\fi
\ifCLASSINFOpdf
\else
\fi

\usepackage{stfloats}
\begin{document}
%
\title{Multi-Order Networks for Action Unit Detection}
%
%
%
%

\author{Gauthier ~Tallec, 
        Arnaud ~Dapogny, 
        and Kévin ~Bailly} 

\IEEEtitleabstractindextext{%
\begin{abstract}
    \tac{Action Units (AU) are muscular activations used to describe facial expressions. Therefore accurate AU recognition unlocks unbiaised face representation which can improve face-based affective computing applications. From a learning standpoint AU detection is a multi-task problem with strong inter-task dependencies. To solve such problem, most approaches either rely on weight sharing, or add explicit dependency modelling by decomposing the joint task distribution using Bayes chain rule. If the latter strategy yields comprehensive inter-task relationships modelling, it requires imposing an arbitrary order into an unordered task set. Crucially, this ordering choice has been identified as a source of performance variations. In this paper, we present Multi-Order Network (MONET), a multi-task method with joint task order optimization. MONET uses a differentiable order selection to jointly learn task-wise modules with their optimal chaining order. Furthermore, we introduce warmup and order dropout to enhance order selection by encouraging order exploration. Experimentally, we first demonstrate MONET capacity to retrieve the optimal order in a toy environment. Second, we validate MONET architecture by showing that MONET outperforms existing multi-task baselines on multiple attribute detection problems chosen for their wide range of dependency settings. More importantly, we demonstrate that MONET significantly extends state-of-the-art performance in AU detection.}
\end{abstract}

}

\maketitle

\IEEEdisplaynontitleabstractindextext

%
\IEEEpeerreviewmaketitle

\IEEEraisesectionheading{\section{Introduction}\label{sec:introduction}}

\tac{Facial expressions are the main channel human uses to convey nonverbal emotional information. Consequently, efficient facial expression detection is key to better face-based computational emotion representation and human-machine interaction. However, such detection performance is limited by the person-specific aspect of facial expressions. To remove person-specific biases, the Facial Action Coding System (FACS) anatomically describe face expressions using a set of unitary muscular activations called Action Units (AU) \cite{ekman1997face}.}

\tac{Using FACS, facial expression detection is equivalent to the joint detection of each considered AUs. From a machine learning point of view, it can be framed as a multi-task problem in which each task corresponds to the prediction of a single AU.}

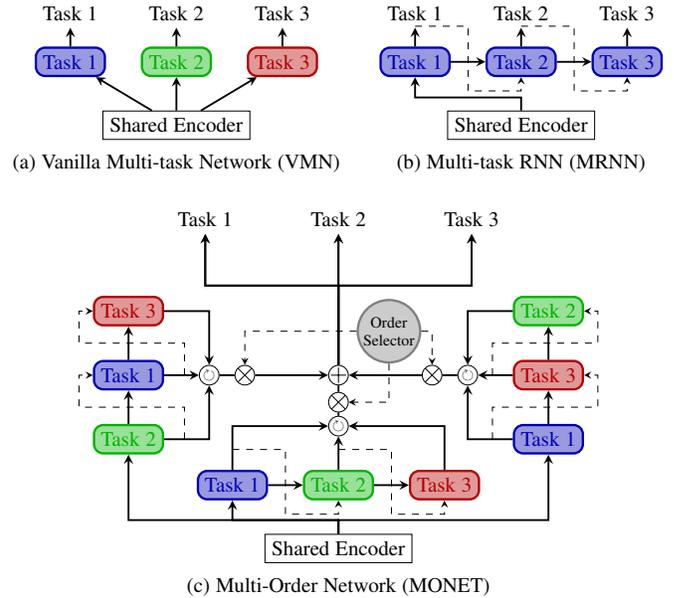
\begin{figure}
\vspace{-0.5cm}
	\resizebox{\columnwidth}{!}{
	\begin{tikzpicture}

	\node[rectangle, draw] (venc) {Shared Encoder};
	\node[below=0.1\vertspace of venc] {(a) Vanilla Multi-task Network (VMN)};
	\node[rectangle, draw, text=black!20!green, above=0.5\vertspace of venc, task2] (vtask2) {Task 2};
	\node[rectangle, draw, text=black!15!blue, left=0.5\horspace of vtask2, task1] (vtask1) {Task 1};
	\node[rectangle, draw, text=black!15!red, right=0.5\horspace of vtask2, task3] (vtask3) {Task 3};

	\coordinate[above=0.5\vertspace of vtask1] (vtask1sup);
	\node at (vtask1sup) (vtask1top) {Task 1};

	\coordinate[above=0.5\vertspace of vtask2] (vtask2sup);
	\node at (vtask2sup) (vtask2top) {Task 2};

	\coordinate[above=0.5\vertspace of vtask3] (vtask3sup);
	\node at (vtask3sup) (vtask3top) {Task 3};

	\draw[-stealth, thick] (venc) -- (vtask1);
	\draw[-stealth, thick] (venc) -- (vtask2);
	\draw[-stealth, thick] (venc) -- (vtask3);

	\draw[-stealth, thick] (vtask1) -- (vtask1top);
	\draw[-stealth, thick] (vtask2) -- (vtask2top);
	\draw[-stealth, thick] (vtask3) -- (vtask3top);

	\node[rectangle, draw, right=\horspace of venc] (renc) {Shared Encoder};
	\node[below=0.1\vertspace of renc] {(b) Multi-task RNN (MRNN)};

	\node[rectangle, draw, text=black!15!blue, above=0.5\vertspace of renc, task1] (rtask2) {Task 2};
	\node[rectangle, draw, text=black!15!blue, left=0.5\horspace of rtask2, task1] (rtask1) {Task 1};
	\node[rectangle, draw, text=black!15!blue, right=0.5\horspace of rtask2, task1] (rtask3) {Task 3};

	\coordinate[below=0.3\vertspace of rtask1] (rtask1inf);
	\coordinate[below=0.2\vertspace of rtask2] (rtask2inf);
	\coordinate[below=0.2\vertspace of rtask3] (rtask3inf);

	\coordinate[above=0.3\vertspace of rtask1] (rtask1infsup);
	\coordinate[above=0.3\vertspace of rtask2] (rtask2infsup);
	\coordinate[above=0.3\vertspace of rtask3] (rtask3infsup);

	\coordinate[above=0.5\vertspace of rtask1] (rtask1sup);
	\node at (rtask1sup) (rtask1top) {Task 1};

	\coordinate[above=0.5\vertspace of rtask2] (rtask2sup);
	\node at (rtask2sup) (rtask2top) {Task 2};

	\coordinate[above=0.5\vertspace of rtask3] (rtask3sup);
	\node at (rtask3sup) (rtask3top) {Task 3};

	\draw[-stealth, thick] (renc) |- (rtask1inf) -- (rtask1);
	\draw[-stealth, thick] (rtask1) -- (rtask2);
	\draw[-stealth, thick] (rtask2) -- (rtask3);

	\draw[-stealth, dashed] (rtask1infsup) -| ($(rtask1) !.5! (rtask2)$) |- (rtask2inf) -- (rtask2);
	\draw[-stealth, dashed] (rtask2infsup) -| ($(rtask2) !.5! (rtask3)$) |- (rtask3inf) -- (rtask3);

	\draw[-stealth, thick] (rtask1) -- (rtask1top);
	\draw[-stealth, thick] (rtask2) -- (rtask2top);
	\draw[-stealth, thick] (rtask3) -- (rtask3top);

	\node[rectangle, draw, below left=6.0\vertspace and 0.5\horspace of renc] (penc) {Shared Encoder};
	\node[below=0.1\vertspace of penc] {(c) Multi-Order Network (MONET)};
	\coordinate[above=0.2\vertspace of penc] (parrow);

	\node[rectangle, draw, text=black!20!green, above=0.5\vertspace of penc, task2] (p12) {Task 2};
	\node[rectangle, draw, text=black!15!blue, left=0.5\horspace of p12, task1] (p11) {Task 1};
	\node[rectangle, draw, text=black!15!red, right=0.5\horspace of p12, task3] (p13) {Task 3};

	\coordinate[below=0.3\vertspace of p11] (p11inf);
	\coordinate[below=0.2\vertspace of p12] (p12inf);
	\coordinate[below=0.2\vertspace of p13] (p13inf);

	\coordinate[above=0.3\vertspace of p11] (p11infsup);
	\coordinate[above=0.3\vertspace of p12] (p12infsup);
	\coordinate[above=0.3\vertspace of p13] (p13infsup);

	\node[circle, draw, above=0.5\vertspace of p12] (p1switch) {};
	\node[scale=0.75] at (p1switch) {$\circlearrowright$}; 
	\node[circle, draw, cross, above=0.05\vertspace of p1switch] (p1mult) {}; 

	\draw[-stealth, thick] (penc) -- (parrow) -| (p11);
	\draw[-stealth, thick] (p11) -- (p12);
	\draw[-stealth, thick] (p12) -- (p13);

	\draw[-stealth, dashed] (p11infsup) -| ($(p11) !.5! (p12)$) |- (p12inf) -- (p12);
	\draw[-stealth, dashed] (p12infsup) -| ($(p12) !.5! (p13)$) |- (p13inf) -- (p13);

	\draw[-stealth, thick] (p11) |- (p1switch);
	\draw[-stealth, thick] (p12) -- (p1switch);
	\draw[-stealth, thick] (p13) |- (p1switch);


	\node[rectangle, draw, text=black!20!green, above left=1.2\vertspace and 1.5\horspace of penc, task2] (p21) {Task 2};
	\node[rectangle, draw, text=black!20!red, above=0.5\vertspace of p21, task1] (p22) {Task 1};
	\node[rectangle, draw, text=black!20!blue, above=0.5\vertspace of p22, task3] (p23) {Task 3};

	\coordinate[left=0.3\vertspace of p21] (p21inf);
	\coordinate[below=0.3\vertspace of p21] (p21below);
	\coordinate[left=0.2\vertspace of p22] (p22inf);
	\coordinate[left=0.2\vertspace of p23] (p23inf);

	\coordinate[right=0.3\vertspace of p21] (p21infsup);
	\coordinate[right=0.3\vertspace of p22] (p22infsup);
	\coordinate[right=0.3\vertspace of p23] (p23infsup);

	\node[circle, draw, right=0.5\horspace of p22] (p2switch) {};
	\node[scale=0.75] at (p2switch) {$\circlearrowright$};
	\node[circle, draw, cross, right=0.2\horspace of p2switch] (p2mult) {}; 

	\draw[-stealth, thick] (penc) -- (parrow) -| (p21);
	\draw[-stealth, thick] (p21) -- (p22);
	\draw[-stealth, thick] (p22) -- (p23);

	\draw[-stealth, dashed] (p21infsup) |- ($(p21) !.5! (p22)$) -| (p22inf) -- (p22);
	\draw[-stealth, dashed] (p22infsup) |- ($(p22) !.5! (p23)$) -| (p23inf) -- (p23);

	\draw[-stealth, thick] (p21) -| (p2switch);
	\draw[-stealth, thick] (p22) -- (p2switch);
	\draw[-stealth, thick] (p23) -| (p2switch);


	\node[rectangle, draw, text=black!20!blue, above right=1.2\vertspace and 1.5\horspace of penc, task1] (p31) {Task 1};
	\node[rectangle, draw, text=black!15!red, above=0.5\vertspace of p31, task3] (p32) {Task 3};
	\node[rectangle, draw, text=black!20!green, above=0.5\vertspace of p32, task2] (p33) {Task 2};

	\coordinate[right=0.3\vertspace of p31] (p31inf);
	\coordinate[below=0.3\vertspace of p31] (p31below);
	\coordinate[right=0.2\vertspace of p32] (p32inf);
	\coordinate[right=0.2\vertspace of p33] (p33inf);

	\coordinate[left=0.3\vertspace of p31] (p31infsup);
	\coordinate[left=0.3\vertspace of p32] (p32infsup);
	\coordinate[left=0.3\vertspace of p33] (p33infsup);

	\node[circle, draw, left=0.5\horspace of p32] (p3switch) {};
	\node[scale=0.75] at (p3switch) {$\circlearrowright$};
	\node[circle, draw, cross, left=0.2\horspace of p3switch] (p3mult) {}; 

	\draw[-stealth, thick] (penc) -- (parrow) -| (p31);
	\draw[-stealth, thick] (p31) -- (p32);
	\draw[-stealth, thick] (p32) -- (p33);

	\draw[-stealth, dashed] (p31infsup) |- ($(p31) !.5! (p32)$) -| (p32inf) -- (p32);
	\draw[-stealth, dashed] (p32infsup) |- ($(p32) !.5! (p33)$) -| (p33inf) -- (p33);

	\draw[-stealth, thick] (p31) -| (p3switch);
	\draw[-stealth, thick] (p32) -- (p3switch);
	\draw[-stealth, thick] (p33) -| (p3switch);

	\node[circle, draw, plus] at ($(p2mult) !.5! (p3mult)$) (plus) {};

	\draw[thick] (p1switch) -- (p1mult);
	\draw[thick] (p1mult) -- (plus);

	\draw[thick] (p2switch) -- (p2mult);
	\draw[-stealth, thick] (p2mult) -- (plus);

	\draw[thick] (p3switch) -- (p3mult);
	\draw[-stealth, thick] (p3mult) -- (plus);

	\coordinate[above=1.2\vertspace of plus] (topsplit);
	\coordinate[above left=1.0\vertspace and 2.0\horspace of topsplit] (leftpred);
	\coordinate[above right=1.0\vertspace and 2.0\horspace of topsplit] (rightpred);
	\coordinate[above=1.0\vertspace of topsplit] (middlepred);

	\node[draw, circle, fill=gray, fill opacity=0.4, line width=1pt, color=gray,  minimum width=3.0em, above right=0.2\vertspace and 0.3\vertspace of plus] (mixture) {};
	\node[scale=0.75, align=center] at (mixture) {Order \\ Selector};

	\node at (leftpred) (task1) {Task 1};
	\node at (middlepred) (task2) {Task 2};
	\node at (rightpred) (task3) {Task 3};

	\draw[-stealth, thick] (plus) -- (topsplit) -| (task1);
	\draw[-stealth, thick] (plus) -- (topsplit) -| (task2);	
	\draw[-stealth, thick] (plus) -- (topsplit) -| (task3);

	\draw[-stealth, thick] (plus) -- (topsplit) -| (task1);
	\draw[-stealth, thick] (plus) -- (topsplit) -| (task2);	
	\draw[-stealth, thick] (plus) -- (topsplit) -| (task3);

	\draw[-stealth, dashed] (mixture) -| (p3mult);
	\draw[-stealth, dashed] (mixture) |- (p1mult);
	\draw[-stealth, dashed] (mixture) -| (p2mult);

\end{tikzpicture}}
	\caption{Overview of MONET along with existing multi-task methods. \tac{Modules (rounded rectangles) with the same color share weights.} (a) VMN models inter-task relationships through sharing encoder weights. (b) MRNN imposes an arbitrary task order. (c) MONET predicts the set of tasks in different orders using separate recurrent cells, one for each task. Those predictions are normalized to a common order (rounded arrows) so that they can be weighted by a convex combination to output the final prediction. By jointly learning the coefficients associated with each order along with the task-specific cells, MONET smoothly selects the best order of prediction and achieves superior performance.
	}
	\label{fig:monet_overview}
	\vspace{-0.5cm}
\end{figure}

\tac{The most widely adopted strategy \cite{ranjan2017hyperface, kokkinos2017ubernet} for solving  multi-task problems is to use a common encoder and predicts the different tasks using separate regressors in parallel (Figure \ref{fig:monet_overview}-(a)). Yet, this strategy fails to model inter-task a priori dependencies. This is all the more a problem as AU are known to display such strong inter-task dependencies. For example, One cannot simultaneously raise and frown his/her eyebrows, so that AU1 and AU4 cannot co-occur.} 

\tac{To better model inter-task dependencies, several works \cite{nam2017maximizing,liu2016recurrent} leveraged recurrent architectures that predict tasks in a sequential fashion (Figure \ref{fig:monet_overview}-(b)). However, contrary to standard sequence processing, the set of tasks to predict has no natural order \textit{a priori}. Therefore, making task prediction sequential requires enforcing an arbitrary order into the set of tasks.
Most importantly, recurrent networks performance have been proven sensitive to the order in which elements are predicted \cite{vinyals2015order}. Consequently, in the frame of multi-task learning using recurrent networks where predicted elements are tasks, the task prediction \textit{order matters}.}

In this paper, we follow this sequential order optimization paradigm and introduce Multi-Order Network (MONET). MONET leverages permutation matrices to represent task orders. More precisely, it navigates between orders by learning a convex combination of permutation matrices (a soft order). In that extent MONET explores the convex hull of permutation matrices (Birkhoff's polytope) to smoothly select the best order. Figure \ref{fig:monet_overview}-(c) illustrates MONET architecture for a 3-tasks problem. To summarize, the contributions of this work are the following:

\begin{itemize}
	\item We introduce MONET, a multi-task learning method with joint task order optimization and prediction. From an architectural standpoint, MONET uses soft order selection in Birkhoff's polytope as well as task-wise cell sharing to model task order and prediction in an end-to-end manner.
	\item From a learning standpoint, we introduce order dropout and warm up strategies that work hand in hand with the order selection to encourage modules to keep good predictive performances in several orders of prediction.
	\item Experimentally, \tac{we first validate MONET architecture in controlled environments by a)Verifying MONET capacity to retrieve the correct order on a toy dataset b) Showing that MONET outperforms several multi-task approaches on a wide range of attribute detection problems with diverse levels of inter-task dependency }. Finally, we conclude by demonstrating that MONET extends state-of-the-art performance in action unit detection.
\end{itemize}

\section{Related Work}
\label{sec:related}
\subsection{Action Unit Detection.}

From a learning perspective, the action unit detection comes with three main specifities. First, action units are local events meaning that the face zones in which they occur are relatively small and constrained. For instance AU1 (inner brow raiser) and AU2 (outer brow raiser) can only occur on the forehead. To tackle that challenge, prior work \cite{zhao2016deep} relaxed convolutional layers weight sharing constraint by introducing region-wise filters. The incentive is that using different filters on each region will result in a refined skin texture representation. However, the performance of such method is conditioned upon face alignment i.e upon weither or not specific face parts are always located in the same rectangular region. To circumvent this limitation several works \cite{li2017eac, shao2018deep, shao2020jaa} used facial key-points along with predefined attention maps to guide the extraction of AU-related features. More precisely, in  \cite{li2017eac}, predefined key-points were used to crop AU related zones while the works in \cite{li2017eac} and \cite{shao2019facial} used attention-like mechanisms based on fixed and jointly learned facial landmark predictors respectively.

Second, because of both social and physiological constraints, action units display strong dependencies between each other (eg: AU1 and AU2 often co-occurs while AU1 and AU4 are incompatible). From an intuitive point of view the exploitation of those dependencies is key to unlock better performance. Indeed, it could enable the use of easy-to-predict AU as a proxy to estimate the harder ones. For example, predicting first AU4 (brow lowerer) could help predict more subtle movements of the eyes such as AU1 or AU6 (cheek raiser). Several works focus on explicitly modeling those dependencies. In particular, backpropagation through a probabilistic graphical model (PGM) was adopted in \cite{Corneanu_2018_ECCV}, an hybrid message passing strategy was used in \cite{song2021hybrid}, \tac{and a Graph Neural Network (GNN) was employed in \cite{li2019semantic}}. Others leveraged the local nature of AU to assume that label dependencies imply dependencies between local face zones. Attempts at capturing such spatial dependencies include attention map learning \cite{shao2019facial}, LSTM-based spatial pooling \cite{niu2019local} and more recently transformer-like architectures \cite{jacob2021facial}.

\tac{Finally, the Action Units annotation work is time consuming and requires specifically trained experts. Hence the lack of a large scale AU annotated database. A way to bypass this issue is to help the AU detection learning with categorical facial expression recognition (FER) labels (happiness, sadness, surprise, fear, disgust) which are more easily obtained. Following this idea, multiple works \cite{cui2020knowledge, wang2017expression} leveraged EMFACS \cite{friesen1983emfacs} a priori dependency to learn to predict AU labels from a pretrained network facial expression predictions. Concurrently, the AU/FER joint learning have been tackled from a multi-task learning perspective: In \cite{wang2019multi}, a cross-stitch-like architecture \cite{misra2016cross} is used to learn to share features between two networks trained on AU and FER datasets respectively. Similarly, in \cite{li2021meta}, a single network is trained on both AU and FER datasets using an adaptative loss weighting. However the benefits of such approaches are limitated by the absence of a dataset with joint AU and FER annotations which prevent AU/FER dependency learning.}


\subsection{Multi-Task Learning}
Deep multi-task learning methods is a subset of deep learning methods that aim at exploiting similarities between several tasks in order to improve individual task performance. To reach that goal, the most widely adopted method relies on implicit modelling of task dependencies using weight sharing. In a nutshell, it consists in splitting the model into parts that are shared across tasks and parts that are task specific.

Seminal work \cite{ranjan2017hyperface, zhang2014facial, wu2015deep, kokkinos2017ubernet}, adopted this weight sharing strategy by making use of a common encoder along with task specific regressors. The intuition behind this is that forcing the same learned features to predict several related task should encourage the encoder to produce more general representations and consequently improve generalization performances. However, one weakness of this method comes with deciding how far features should be shared. Indeed, it intuitively depends on task relatedness (the more related the tasks the further the sharing) which can be hard to determine. To tackle this issue, numerous approaches \cite{ruder2017overview, misra2016cross, gao2019nddr, LiuSY19} used adaptative architectures to jointly learn which layers should be shared between tasks, as well as the task prediction itself. This philosophy has also been used in modular approaches which consist in learning a set of trainable modules along with how they should be combined for each tasks. For instance, soft layer ordering \cite{meyerson2017beyond} consists in learn the best module combination for each task in a fully differentiable way. In the same vein, a select or skip policy \cite{sun2020adashare} was used to determine which module should be used for each task.

 Weight sharing may help finding features that are useful for all tasks and therefore implicitly models input-related task conditional dependencies, though, it doesn't capture inter-task relationships that do not depend on input (e.g the prior that detection of a beard implies high probability that the subject also has a mustache). In order to model those dependencies, several approaches \cite{ruder2017overview, nam2017maximizing, bilen2016integrated} leveraged recurrent neural networks to decompose the task joint distribution into a product of conditional distributions using Bayes chain rule. Most importantly, the work in \cite{vinyals2015order} showed that \textit{order matters}, meaning that the order in which the chain rule is unrolled impacts the final joint estimate modelization performance. In the light of this observation it proposes a two-steps method: The first step consists in an exploration phase in which the performance of several orders are tested. At the end of this phase, a single order is fixed once and for all based on the exploration phase performance and predictions are computed using this order.

Our work lies in the continuity of the order optimization paradigm proposed in \cite{vinyals2015order} which aim at improving current Bayes chain rule based joint distribution estimation. However, we stand out from it by drawing inspiration in \cite{meyerson2017beyond} to: (a) Propose a soft order selection mechanism that navigates through Birkhoff's polytope, and (b) Propose a new task-wise modular recurrent architectural design. More precisely, MONET smooth selection contrasts with the once and for all choice of order in \cite{vinyals2015order} by keeping on learning several orders during all the training phase. We take full advantage of this by adding warm up and order dropout mechanisms that encourage modules to display good predictive performances for several orders of prediction.

By optimizing the task order, MONET takes advantage of situations where \textit{order matters}. Furthermore, we believe that learning more than one task order all along the training improves MONET generalization capacity and thus predictive performance in more general multi-task settings where order do not necessarily matter.

\section{Methodology}
\label{sec:methodology}

In the rest of the paper, scalars are denoted using regular characters, vectors are in bold. For vector $\mathbf{v}$ the t-th coordinate is denoted by $v^{(t)}$, and the vector of its $t-1$ first coordinates is denoted by $\mathbf{v}^{(<t)}$. 
$\mathcal{D} = \{(\mathbf{x}_{i}, \mathbf{y}_{i})\}_{i=1}^{N}$ is a training dataset composed of input vectors $\mathbf{x}$ and labels $\mathbf{y}$ of size $T$ such that $\forall t \in [1, T], y^{(t)} \in \{0, 1\}$.


\subsection{Multi-task baselines}
\subsubsection{Vanilla multi-task Networks (VMN)}
The most widely adopted deep multi-task approach is to model task dependencies using weight sharing only \cite{dong2015multi,sogaard2016deep}, i.e to assume labels conditional independance given the input image. Template networks for this approach are composed of a shared encoder $f_{\mathbf{W}}$ parametrized by matrix $\mathbf{W}$ along with a specific prediction head $g_{\mathbf{W}^{(t)}}$, parametrized by $\mathbf{W}^{(t)}$ for each task $t \in [1, T]$. We refer to instances of such template as Vanilla Multi-task Networks (VMN). Given input $\mathbf{x}$, the prediction for task $t$ is the output of the $t$-th prediction head: 
\begin{equation}
    p^{(t)} = S(g_{\mathbf{W}^{(t)}} \circ f_{\mathbf{W}}(\mathbf{x})),
\end{equation}
where $S$ denotes the sigmoid function. Task $t$ distribution is then estimated as follows:
\begin{equation}
\begin{aligned}
    \log p_{\theta}(y^{(t)} \mid \mathbf{x}) &= - \text{BCE}(y^{(t)}, p^{(t)}), \\
                                             &= y^{(t)} \log p^{(t)} + (1 - y^{(t)}) \log (1 - p^{(t)}),
\end{aligned}
\end{equation}
where $\theta = \{\mathbf{W}, (\mathbf{W}^{(t)})_t\}$, and BCE stands for binary cross entropy. Training is done by minimizing the following maximum likelihood-based loss: 
\begin{equation}
	\mathcal{L}(\theta) = - \sum\limits_{i=1}^{N}\sum\limits_{t=1}^{T}\log p_{\theta}(y^{(t)}_i \mid \mathbf{x}_i).\\
\end{equation}

If VMN are the most classic multitask learning approach, their modelization assumption do not take into account inter-task relationships that are independent of the input image. Those relationships include numerous human knowledge-based priors such as the statistical dependency between the presence of a beard and the presence of a mustache, for instance. Naturally, predictive performance of a deep network may benefit from exploiting these priors.

\subsubsection{Multi-task Recurrent Neural Networks (MRNN)}

Different from VMN, Multi-task Recurrent Neural Networks (MRNN) model inter-task dependencies through both weight sharing and joint conditional distribution modelization. For that purpose, the joint conditional distribution of labels is decomposed using Bayes chain rule:
\begin{equation}
	p(\mathbf{y} \mid \mathbf{x}) = \prod_{t=1}^{T} p(y^{(t)} \mid \mathbf{y}^{(<t)}, \mathbf{x}).
	\label{eq:chain_rule}
\end{equation}

The MRNN approach consists in encoding the input vectors with network $f_{\mathbf{W}}$ and to feed the output representation as the initial state $\mathbf{h}^{(0)}$ of a recurrent computation process driven by cell $g_\mathbf{V}$ with parameters $\mathbf{V}$. At step $t$, this process takes one hot encoded ground truth for timestep $t - 1$ task along with hidden state $\mathbf{h}^{(t - 1)}$ and outputs prediction $p^{(t)}$ and next timestep hidden state $\mathbf{h}^{(t)}$. In a nutshell:
\begin{equation}
\begin{aligned}
	\mathbf{h}^{(0)} &= f_{\mathbf{W}}(\mathbf{x}), y^{(0)} = 0,\\
	o^{(t)}, \mathbf{h}^{(t)} &= g_{\mathbf{V}}(\mathbf{h}^{(t-1)}, \tilde{y}^{(t-1)}\mathbf{e}_{t-1}).
\end{aligned}
\end{equation}
where $\tilde{y}^{(t-1)} = 2(y^{(t-1)} - 1)$ and $(\mathbf{e}_{1}, \dots, \mathbf{e}_{T})$ denotes the canonical basis vectors of $\mathbb{R}^T$.
Prediction $p^{(t)}$ is then: 
\begin{equation}
	p^{(t)} = S(o^{(t)}).
\end{equation}
Task $t$ conditional distribution w.r.t previous tasks is estimated as:
\begin{equation}
    \log p_{\theta}(y^{(t)} \mid \mathbf{y}^{(<t)}, \mathbf{x}) = - \text{BCE}(y^{(t)}, p^{(t)}),
\end{equation}
where $\theta = \{\mathbf{V}, \mathbf{W}\}$, and training is done by minimizing the following loss: 
\begin{equation}
	\mathcal{L}(\theta) = -\sum\limits_{i=1}^{N}\sum\limits_{t=1}^{T}\log p_{\theta}(y_i^{(t)} \mid \mathbf{y}_{i}^{(<t)}, \mathbf{x}_{i}).
\end{equation}

For task $t$ and input $\mathbf{x}$, inference consists in estimating $p_{\theta^{*}}(y^{(t)} \mid \mathbf{x})$, where $\theta^{*} = \{\mathbf{W}^{*}, \mathbf{V}^{*}\}$ denotes MRNN parameters at the end of the training phase. To compute this estimation we leverage Monte-Carlo sampling in the following way:
\begin{equation}
\begin{aligned}
p_{\theta^{*}}(y^{(t)} \mid \mathbf{x}) &= \mathbb{E}_{\mathbf{y}^{(<t)}}[p_{\theta^{*}}(y^{(t)} \mid \mathbf{y}^{(<t)}, \mathbf{x})],\\
& \simeq \frac{1}{L} \sum\limits_{l=1}^{L} p_{\theta^{*}}(y^{(t)} \mid \hat{\mathbf{y}}_{l}^{(<t)}, \mathbf{x}),
\label{eq:MRNN_sampling}
\end{aligned}
\end{equation}
where $(\hat{\mathbf{y}}_{l})$ for sample $l\in[1, L]$ are computed sequentially, as summarized in algorithm \ref{algo:MRNN_sampling}.

\begin{algorithm}[H]
\caption{MRNN sampling for inference \label{algo:MRNN_sampling}}
\begin{algorithmic}[1]
\REQUIRE Input vector $\mathbf{x}$
  \FOR{$l=1$ \TO $L$}
    \STATE $\hat{y}^{(0)}_{l} = 0$, $\mathbf{h}_{l}^{(0)}=f_{\mathbf{W^{*}}}(\mathbf{x})$
    \FOR{ $t=1$ \TO $T$ }
        \STATE $o_{l}^{(t)}, \mathbf{h}_{l}^{(t)} = g_\mathbf{V^{*}}(\mathbf{h}_l^{(t-1)}, \tilde{y}_{l}^{(t-1)}\mathbf{e}_{t-1})$
        \STATE $p_l^{(t)} = S(o_l^{(t)})$
        \STATE $\hat{y}_l^{(t)} \sim \mathcal{B}(p_{l}^{(t)})$
   \ENDFOR
  \ENDFOR
\RETURN $L$ output trajectories $(\hat{\mathbf{y}}_{l})$
\end{algorithmic}
\end{algorithm}

The performances of MRNN are directly linked to the conditional joint distribution estimate modelization performance that itself depends on the modelization performance of each element in the chain rule product.
In \cite{vinyals2015order}, it is established that \textit{order matters}, meaning that unrolling chain rule in a different task chaining order leads to different modelization performance. This comes from the fact that tasks may be easier to learn in a given order. By relying on a single arbitrary chain rule decomposition order, MRNN misses the opportunity to better exploit the inter-task relationships. By contrast, our method extends MRNN by parallely estimating the joint conditional distribution using different orders and smoothly selecting the best estimate. 

\subsection{Multi-Order Network (MONET)}

\begin{figure*}
\centering
\begin{adjustbox}{width=\linewidth}
\input{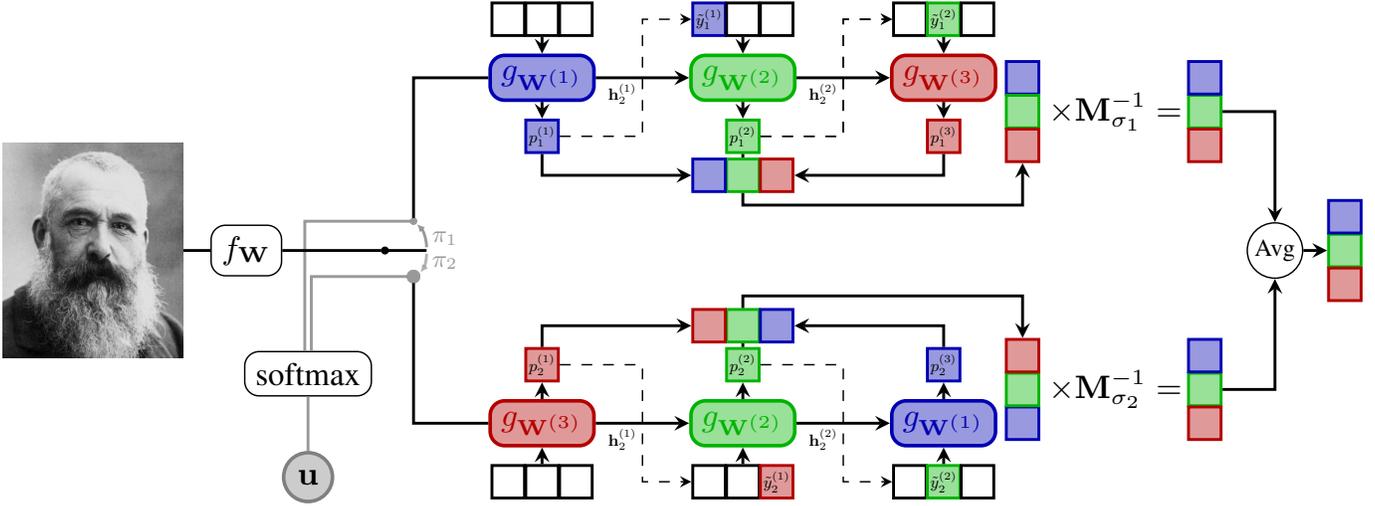}
\end{adjustbox}
\caption{MONET inference overview with $T=3$ tasks and $M=2$ orders $\sigma_1 = [1, 2, 3]$ and $\sigma_2 = [3, 2, 1]$. \tac{All visual elements with the same color are relative to the same task (Blue for task 1, green for task 2 and red for task 3), rounded rectangles represent prediction modules while straight corner rectangles represent their predictions}.  At inference time, MONET predicts all tasks in $L$ different orders sampled from order selector $\pi$. The final prediction is then the average of those $L$ predictions. Consequently, the tasks are predicted in orders that are learned at train time. Hence it better captures inter-task dependencies and yields enhanced multi-task performance. }
\label{monet2}
\end{figure*}

In this section, we present MONET (Multi-Order Network) for joint task order and prediction modelling in multi-task learning. MONET is composed of an order selector that navigates through Birkhoff's polytope to learn a suitable task order in a differentiable way. As illustrated on Figure \ref{monet2}, in inference mode, sampling from this order selector allows MONET to combine its $T$ recurrent cells, one for each task, in an order that has been learned at train time.

\subsubsection{Jointly learning task order and prediction}

Let's define a soft order of $T$ tasks as any real doubly stochastic matrix $\Omega$ of size $T \times T$, \textit{i.e.} a matrix such that :
\begin{equation}
    \forall i,j \in [1,T]:  \sum_{k=1}^T \Omega_{i,k} = \sum_{k=1}^T \Omega_{k,j}=1.   
\end{equation}

 Intuitively, in such case, the coefficient $\Omega_{i,j}$ associated to each row $i$ and column $j$ in $\Omega$ corresponds to the probability to address task $j$ at step $i$. Therefore,  in the extreme situation where all columns are one-hot vectors, a soft order matrix becomes a ``hard'' order (\textit{i.e.}, a permutation matrix) that models a deterministic task order. More precisely, if $\sigma$ denotes a permutation, its associated order matrix is :
\begin{equation}
	(\mathbf{M}_{\sigma})_{i, j} = 
\begin{cases}
	1 &\text{ if }j = \sigma(i), \\
	0 &\text{ otherwise. }
\end{cases} 
\end{equation}

The Birkhoff-Von Neumann's theorem states that the class of doubly stochastic matrices (also called Birkhoff's polytope) is the convex hull of all the order matrices, \tac{i.e the set of all convex combinations of order matrices}. In other words, any soft order matrix $\Omega$ can be decomposed as a convex combination of $M$ order matrices. Formally, there exists $M$ a finite number, $\pi_1, \dots, \pi_M$ $\in \mathbb{R}$, and $\mathbf{M}_{\sigma_1}, \dots, \mathbf{M}_{\sigma_M}$, $M$ order matrices such that:
\begin{equation}
	\Omega = \sum_{m=1}^{M}\pi_m\mathbf{M}_{\sigma_m}.
\end{equation}

Therefore, each soft order of $T$ tasks can be parametrized by the coefficients $\pi_1, \dots \pi_M$ associated to each possible order matrices, with $M=T!$. The reciprocal is also true: given $M$ order matrices, with $M \leq T!$, each convex combination $\pi_1, \dots, \pi_M$ also defines a soft order.

We use this result to provide a differentiable parametrization of soft orders, that allows us to jointly learn both the task order and prediction by smoothly navigating Birkhoff's polytope. To do so, we first generate $M \leq T!$ random permutations, denoted as $(\sigma_m)_{m \in [1,M]}$. For each permutation, $\sigma$, we estimate a joint distribution convex $p^{\sigma}$ by unrolling the Chain Rule in order $\sigma$:
\begin{equation}
	p^{\sigma}(\mathbf{y} \mid \mathbf{x}) = \prod_{t=1}^{T} p^{\sigma}(y^{\sigma(t)} \mid \mathbf{y}^{\sigma(<t)}, \mathbf{x})
	\label{eq:chain_rule_extended}
\end{equation}

Finally, we compute the final joint distribution as a convex combination of each permutation-based joint distribution:
\begin{equation}
	p(\mathbf{y} \mid \mathbf{x}) = \sum\limits_{m=1}^{M} \pi_m p^{\sigma_m}(y^{\sigma_m(t)} \mid \mathbf{y}^{\sigma_m(<t)}, \mathbf{x}),
	\label{eq:chain_rule_extended}
\end{equation}
where $\pi_1, \dots, \pi_M$ are the order selector coefficients that position the learned soft order $\Omega = \sum_{m=1}^{M}\pi_m\mathbf{M}_{\sigma_m}$ inside Birkhoff's polytope.


\subsubsection{MONET architecture}

MONET is composed of $T$ recurrent cells $(g_{\mathbf{W}^{(t)}})_{1\leq t \leq T}$, each being trained to predict the same task across all orders: For an order $\sigma$, task $\sigma(t)$ is predicted using recurrent cell $g_{\mathbf{W}^{\sigma(t)}}$ and is conditioned on the results of all preceding tasks in the order $\sigma$ (i.e tasks $\sigma(1), \dots, \sigma(t-1)$). The rationale behind this comes from traditionnal RNN usage, where each cell predicts a single task in different contexts. (e.g., RNN-based sentence translation is a repetition of word translations conditioned by the context of neighbouring words).
Here, each task-associated predictor learns to predict the corresponding task in different contexts corresponding to the different orders. 

Concretely, for order $\sigma_m$, our computational graph unfolds as follows: 
\begin{equation}
\begin{aligned}
	\mathbf{h}^{(0)}_{m} &= f_{W}(\mathbf{x}), y_{m}^{(0)} = 0,\\
	o^{(t)}_m, \mathbf{h}^{(t)}_m &= g_{\mathbf{W}^{\sigma_m(t)}}(\mathbf{h}_{m}^{(t-1)}, \tilde{y}_{m}^{(t-1)}\mathbf{e}_{\sigma_m(t-1)}),
\end{aligned}
\end{equation}
where $\tilde{y}_{m}^{(t-1)} = 2(y^{\sigma_m(t-1)} - 1)$. Prediction $p_m^{(t)}$ at timestep $t$ is computed as :
\begin{equation}
p_{m}^{(t)} = S(o_m^{(t)}),
\end{equation}
and is used as follows: 
\begin{equation}
    \log p^{\sigma_m}_{\theta}(y^{\sigma_m(t)} \mid \mathbf{y}^{\sigma_m(<t)}, \mathbf{x}) = -\text{BCE}(y^{\sigma_m(t)}, p_m^{(t)}).
\end{equation}
Parameters $\theta=(\mathbf{W}, (\mathbf{W}^{(t)})_{1 \leq t \leq T})$ as well as the order selector $\pi$, defined as a softmax layer over logits $\mathbf{u}$, are jointly learned through minimizing the following loss:
\begin{equation}
	\mathcal{L}(\theta, \pi) = -\sum\limits_{i=1}^{N} \log \sum_{m=1}^{M} \exp [\log \pi_m + \log p^{\sigma_m}_{\theta}(\mathbf{y}_i \mid \mathbf{x}_i)].
\label{eq:final_loss}
\end{equation}

\subsection{Order Selection Strategy}
In this section, we provide theoretical intuitions about MONET order selection. In particular, we highlight that the raw MONET order selection mechanism tends to allocate weights on the permutation with the lowest loss. For that purpose, we denote by $\mathcal{L}^{\sigma}$, the loss associated with order $\sigma$: 
\begin{equation}
    \mathcal{L}^{\sigma}(\theta) = - \sum_{i=1}^{N}\log p^{\sigma}_{\theta}(\mathbf{y}_{i} \mid \mathbf{x}_{i}).
\end{equation}
We also introduce element-wise losses for both order based and global losses as :
\begin{equation}
    \begin{aligned}
    \mathcal{L}^{\sigma}_{i}(\theta) &= -\log p^{\sigma}_{\theta}(\mathbf{y}_{i} \mid \mathbf{x}_{i}),\\
    \mathcal{L}_{i}(\theta, \pi) &= -\log \sum_{m=1}^{M} \exp(\log \pi_m  -\mathcal{L}_{i}^{\sigma_{m}}(\theta)).
    \end{aligned}
\end{equation}
MONET order selection is based on the variation of $\pi$, which itself is underpinned by gradient updates on order logits $\mathbf{u}$. Consequently, we compute the loss gradient element-wise. More precisely for $i \in [1, N]$, for $m\in[1, M]$:
\begin{equation}
	\frac{\partial{\mathcal{L}_i(\theta, \pi)}}{\partial{u_{m}}} = \pi_{m}(1 -  \exp(\mathcal{L}_{i}(\theta, \pi) - \mathcal{L}_{i}^{\sigma_m}(\theta)),
	\label{eq:ew_gradient}
\end{equation}
and : 
\begin{equation}
\mathcal{L}_{i}^{\sigma_m}(\theta) \leq \mathcal{L}_{i}(\theta, \pi) \iff \frac{\partial{\mathcal{L}_{i}(\theta, \pi)}}{\partial{u_{m}}} \leq 0.
\label{eq:grad_cond}
\end{equation}
The latter equivalence implies that, in the case of element by element loss minimization, coefficient $\pi_m$ increases if the loss associated to its order is inferior to the global loss.  In a more realistic scenario, optimization is performed by batch of size $B$. The gradient of loss $\mathcal{L}$ on a batch is then: 
\begin{equation}
\begin{aligned}
    \frac{\partial{\mathcal{L}_{B}(\theta, \pi)}}{\partial{u_{m}}} &= \sum_{i=1}^{B} \frac{\partial{\mathcal{L}_{i}(\theta, \pi)}}{\partial{u_{m}}},\\
    &= \sum_{i = 1}^{B} \pi_{m}(1 -  \exp(\mathcal{L}_{i}(\theta, \pi) - \mathcal{L}_{i}^{\sigma_m}(\theta))).\\
    \textit{(Jensen)}&\leq \pi_{m}B(1 - \exp(\frac{1}{B}(\mathcal{L}_{B}(\theta, \pi) - \mathcal{L}_{B}^{\sigma_m}(\theta)))
	\label{eq:grad_cond}
\end{aligned}
\end{equation}
It directly follows that: 
\begin{equation}
\mathcal{L}_{B}^{\sigma_m}(\theta) \leq \mathcal{L}_{B}(\theta, \pi) \Rightarrow \frac{\partial{\mathcal{L}_{B}(\theta, \pi)}}{\partial{u_{m}}} \leq 0.
\label{eq:grad_cond}
\end{equation}

Consequently, orders whose losses are the lowest on a batch get positive updates on their order selector coefficients. As a consequence, using raw MONET order selection results in selecting the order whose joint estimation best fits the train set, i.e the order that overfits the most. This observation comes with two different issues:

\begin{figure}
    \centering
    \includegraphics[width=\linewidth]{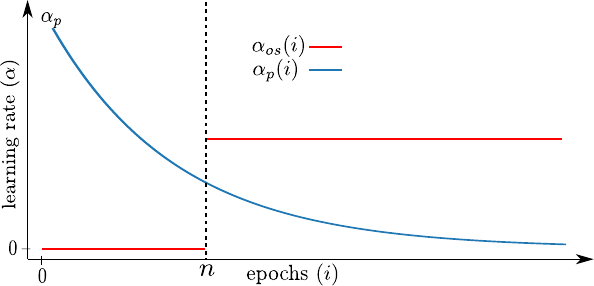}
    \caption{Warm up learning rate schedule. $\alpha_p$ represents the learning rate in the predicting part of the network while $\alpha_{os}$ is the order selector learning rate. The order selector stay frozen for the $n$ first epochs so that the predictor estimates the performance of each orders. At the end of this exploration phase, the order selector coefficients are released and uses those estimations to allocate weight to the orders with the best performance.}
    \label{fig:warmup}
\end{figure}
\subsubsection{Warmup : \tac{stepping away from bad starting points}}
When training begins the order losses mostly depend on network initialisation, rather than on their associated order respective performance. Hence, order selection in the first epochs is likely to lead to quasi-random solutions. This is all the more problematic as weight allocation is prone to snowballing i.e to keep allocating more and more weight to a previously selected order. If, for a given epoch, the loss $\mathcal{L}^{\sigma_i}$ for order $\sigma_i$ is lower than the loss term corresponding to other orders, the proposed method will assign more weight to this order by positively updating the corresponding $\pi_i$. As a consequence, the task modules will specialize in predicting the tasks using order $\sigma_i$. This will reinforce the advantage of order loss $\sigma_i$. This is a problem since, in such a case, the order selection mechanism does never have a chance to really try and compare different orders. To circumvent this, we draw inspiration from the exploration phase in \cite{vinyals2015order} and freeze order logits $\mathbf{u}$ for the first $n$ epochs (see figure \ref{fig:warmup}). This provides the network with an opportunity to explore all orders and to get a better estimate of each order performance. In turn, it improves the quality of the selected orders.

\subsubsection{Order dropout: \tac{avoiding order selection snowball}}
Warm up may help the order selector to choose among the best performing orders. However it doesn't prevent order selection from snowballing on the first order it selects. This snowball effect yields increased risks of getting stuck in a suboptimal order and neglects the benefits of training several orders parallely. To avoid this pitfall, we propose an order dropout strategy which consists in training each example on a random subset of $k$ (Instead of $M$) orders by zeroing-out $(M-k)$ order selector coefficients:
\begin{equation}
	\tilde{\pi}_{m}^{i} = \frac{t_{m}^{i} \exp(u_m)}{\sum\limits_{l=1}^{M} t_{m}^{l} \exp(u_l)},
\end{equation}
where $t^{i}_{m}$ is a randomly sampled binary mask with $k$ ones and $(M-k)$ zeros. For inference, we  multiply each $\exp(u_m)$ by its probability $p(k, M)$ of presence, as in~\cite{dropout}:
\begin{equation}
	\pi_{m} = \frac{p(k, M) \exp(u_m)}{\sum\limits_{l=1}^{M} p(k, M) \exp(u_l)} = \frac{\exp(u_m)}{\sum\limits_{l=1}^{M} \exp(u_l)}.
\end{equation}
 With this strategy the order with the lowest loss is not always included in the $k$ trained order and do not get systematically reinforced by the successive gradient updates. Therefore it short-circuits the order selector snowballing behaviour and forces MONET to spread weight allocation, hence encouraging good predictive performance for several orders. We believe that forcing each task module to learn its associated task in different order have a regularizing effect and reduces overfitting leading to better predictions at inference time.
 
 \subsection{Inference with MONET}

Let $\theta^{*}, \pi^{*}$ be MONET parameters at the end of the training phase. At test time, for an input $\mathbf{x}$, $R$ different orders $(\sigma_{r})_{1\leq r \leq R}$ are sampled from a random variable $\mathbf{S}$ whose discrete distribution is based on the order selector $\pi^{*}$. The input $\mathbf{x}$ is then routed into the $R$ networks corresponding to each order (figure \ref{monet2}). Each of those networks outputs a prediction for task $t$, and those predictions are averaged to form the global network prediction. Formally :
\begin{equation}
\begin{aligned}
	p_{\theta^{*}}(y^{(t)} \mid \mathbf{x}) &= \mathbb{E}_{\mathbf{S}}[p_{\theta^{*}}(y^{(t)} \mid \mathbf{x}, \mathbf{S})],\\
	  &= \mathbb{E}_{\mathbf{S}}[\mathbb{E}_{\mathbf{y}^{\sigma_\mathbf{S}(<\sigma_\mathbf{S}^{-1}(t))}}[p^{\sigma_{\mathbf{S}}}_{\theta^{*}}(y^{(t)} \mid \mathbf{x}, 
	  \mathbf{y}^{\sigma_\mathbf{S}(<\sigma_\mathbf{S}^{-1}(t))})]], \\
	  &\simeq \frac{1}{LR}\sum_{r=1}^{R}\sum_{l=1}^{L} p^{\sigma_{s_r}}_{\theta^{*}}(y^{(t)} | \hat{\mathbf{y}}_{r, l}^{\sigma_{s_r}(<\sigma_{s_r}^{-1}(t))}, \mathbf{x}),
\end{aligned}
\end{equation}
where $(\hat{\mathbf{y}}_{r,l})$ for $r\in[1, R], l \in [1, L]$ are sampled using algorithm \ref{algo:MONET_sampling}.

\begin{algorithm}[H]
 \caption{MONET sampling for inference \label{algo:MONET_sampling}}
 \begin{algorithmic}[1]
 \REQUIRE Input vector $\mathbf{x}$
 \FOR{$r=1$ \TO $R$}
 	\STATE $s_r \sim \mathbf{S}$ 
    \FOR{$l=1$ \TO $L$}
        \STATE $\hat{y}^{(0)}_{r, l} = 0, \mathbf{h}_{r, l}^{(0)} = f_{\mathbf{W}^{*}}(\mathbf{x})$
        \FOR{$t=1$ \TO $T$}
            \STATE $o_{r, l}^{(t)}, \mathbf{h}_{r, l}^{(t)} = g_{\mathbf{W}^{\sigma_r(t)}}(\mathbf{h}_{r, l}^{(t-1)}, \tilde{y}_{r, l}^{(t-1)}\mathbf{e}_{\sigma_{s_r}(t-1)})$
            \STATE $p_{r, l}^{(t)} = S(o_{r, l}^{(t)})$
            \STATE $\hat{y}^{(t)}_{r, l} \sim \mathcal{B}(p_{r, l}^{(t)})$
        \ENDFOR
    \ENDFOR
 \ENDFOR
 \RETURN $LR$ output trajectories $(\hat{\mathbf{y}}_{r, l})$
\end{algorithmic}
\end{algorithm}

In practice, we found out that, when training ends, $\pi^{*}$ is often close to a one-hot vector. Consequently, the sampling of $R$ orders is likely to result in $R$ times the same order. In that scenario, the final prediction becomes the average of $LR$ trajectories generated using the same prediction order. Also, for the sake of simplicity, we regroup all trajectory numbers in variable $L \gets LR$.
\section{Experiments}
\label{sec:experiments}

\subsection{Datasets}

\begin{figure}[t]
\centering
\includegraphics[width=\linewidth]{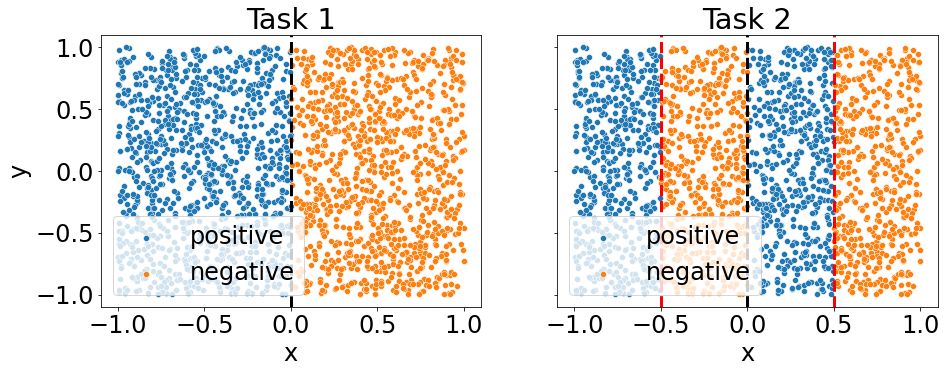}
\caption{Distribution of $2000$ examples sampled from our toy dataset with $T=2$. Given task $1$, i.e given the left-right positioning of the sample w.r.t to the black dashed boundary, task $2$ is simplified into a single linear boundary learning problem represented by red boundaries.
\label{fig:toy_distribution}} 
\end{figure}

\noindent\textbf{Toy Dataset} For method empirical validation, we designed a 2-dimensional multi-task binary classification toy dataset represented in Figure \ref{fig:toy_distribution}. For $T$ tasks, it uses the  following laws for input and labels: 
\begin{equation}
\begin{aligned}
	\mathbf{X} &\sim \mathcal{U}([-1; 1]^{2}),\\
	\forall t\in [1, T] : Y_{T}^{(t)} &= \mathbf{1}_{\bigcup_{i=1}^{2^{t}}[b_{2i}^{(t)}, b_{2i + 1}^{(t)}]}(X^{(1)}),
\end{aligned}
\label{eq:toygen}
\end{equation}
where $b_{i}^{(t)} = -1 + \frac{i-1}{2^{t}}$. In short, $[-1; 1]^{2}$ is vertically split in a recurrent way. This dataset has a natural order in which the tasks are easier to solve: for $t\in [1, T]$, conditioning task $t+1$ by the $t$ first tasks results transforms it into an easy-to-solve classification problem with a single linear boundary. Conversely, conditioning task $t+1$ by the result of upcoming tasks in coordinate order do not simplify its complexity as a classification problem, it remains a $2^{t} - 1$ linear boundary problem . 
We generate 500, 250 and 250 examples for the train, val and test partitions, respectively. Those sizes are deliberately small to challenge networks modelization performance.

\noindent\textbf{CelebA} is a widely used database in multi-task learning, composed of $\approx 200k$ celebrity images annotated with 40 different facial attributes. For performance evaluation, we measure accuracy score using the classic train ($\approx 160k$ images), valid ($\approx 20k$ images) and test ($\approx 20k$ images) partitions for 5 different subsets of 5 attributes each:
\begin{itemize}
    \item \textit{gender}: with moustache, beard, lipstick, heavy makeup and sex detection. Those attributes display statistical dependencies. For example a beard often implies a moustache.
    \item \textit{accessory}: with earrings, eyeglasses, necklaces and neckties detection.
    \item \textit{beauty}: with arched eyebrows, attractiveness , high cheekbones, rosy cheeks, and oval faces detection.
    \item \textit{haircut}: with baldness, black, blond, brown and gray hair detection. Those attributes are mutually exclusive.
    \item \textit{miscellaneous}: with 5 o'clock shadow, pointy nose, mouth slighlty open, oval face, and weither the subject is young or not. Those attributes are independant \textit{a priori}.
\end{itemize}
Both \textit{accessory} and \textit{beauty} were choosen for their lack of clear a priori on the type of dependencies that bind the tasks together. \tac{Those tasks subsets were constructed to assess MONET behaviour in different dependency settings, and show its interest as an overall better multi-task approach compared with existing baselines.}

\noindent\textbf{BP4D} is a dataset for facial action unit detection. It is composed of approximately $140k$ images featuring $41$ people ($23$ female, $18$ male) with different ethnicities. Each image is annotated with the presence of $12$ AU. For performance evaluation, we follow related work strategy that is to report F1-Score on all $12$ AUs using a subject exclusive 3-fold cross-validation with publicly available fold repartition from \cite{shao2018deep} and \cite{shao2020jaa}.

\noindent\textbf{DISFA} is another dataset for facial action unit detection. It contains 27 videos for $\approx 100k$ face images. Those images were collected from $27$ participants and annotated with $12$ AUs. Originally, each AU label is an intensity score ranging from $0$ to $5.$. In detection, labels with an intensity score higher than 2 are considered positive \cite{zhao2016deep}. Similarly to BP4D, the performance evaluation protocol consists in measuring the F1-Score for $8$ AUs using a subject exclusive 3-fold cross validation.


\subsection{Implementation Details}
In our experiments, we compare MONET with several multi-task baselines, each using a shared encoder and a number of prediction heads. For VMN-Common (VMNC), the prediction head consists in two dense $\rightarrow$ BN $\rightarrow$ ReLU applications followed by another dense $\rightarrow$ sigmoid layer with $T$ outputs. VMN-Separate (VMNS) uses $T$ prediction heads, each with the same structure, except the last layer is of size $1$. Finally, MRNN uses a single Gated Recurrent Unit (GRU) cell which sequentially predicts the $T$ tasks. Task order is randomly sampled for each MRNN experiment.

\noindent\textbf{Toy Experiments: } The shared encoder consists of four dense layers with $64$ units and ReLU activation. Prediction heads for both VMNC and VMNS consist in dense layers with $64$ units. Both MRNN and MONET employ GRU cells with $64$ units and $L=20$ orders. All networks are trained by applying 500 epochs with Adam~\cite{adam}, batch size $64$ with an exponentially decaying base learning rate $5e-4$ and $\beta=0.99$.
MONET order selector is trained with Adam with learning rate $0.005$. Other MONET related parameters (dropout $k$, warmup $n$) are determined by hyperparameter tuning on a dedicated validation dataset.

\noindent\textbf{Face Attributes:} For CelebA we make use of an Inception resnet v1 encoder pretrained on VGGFace2 \cite{cao2018vggface2} along with dense layers with $64$ units as prediction heads for VMNC and VMNS. Both MRNN and MONET use GRU cells with $64$ units and $L=20$ orders. Networks are trained with $30$ epochs using AdamW~\cite{adamw} with learning rate/weight decay set to $0.0005$ and exponential decay ($\beta=0.96$). For MONET hyperparameters, we use $M=5!=120$ permutations and order dropout $k=32$.

\noindent\textbf{Facial Action Unit Detection : }
Action Units are relatively short events. Therefore video-based datasets such as DISFA and BP4D display a data imbalance problem \cite{shao2020jaa} that may act as a barrier to efficient learning. To circumvent this problem, we combine  biais initialization \cite{lin2017focal} and per AU loss weighting \cite{shao2020jaa}. To further adapt to the evaluation protocol that measure F1 score we include a Dice score contribution \cite{shao2020jaa, jacob2021facial} to the final loss.

For BP4D, we use an Inceptionv3 backbone pretrained on imagenet on top of which we put a MONET instance with $M=512$ and $k=128$. The order selection part of MONET is trained using Adam with warmup $n=5$ and constant learning rate $5e-3$ while the rest of the network uses AdamW optimizer with learning rate/weight decay set to $1e-4$ and exponential decay $\beta = 0.99$. Batchsize is set to $16$, number of epochs is $2$ and dice loss coefficient is $\lambda=0.5$.

For DISFA, we make use of the same encoder as for CelebA  and retrain with AdamW learning rate $5e-4$ with exponential decay $\beta = 0.96$, and batchsize $64$. All other parameters stay the same as for BP4D.



\subsection{Toy experiments}



\begin{figure}
\centering
    \caption{Performance of MONET trained with a single order $\sigma$ $(M=1, k=1)$ as a fonction of the frobenius distance between $\sigma$ and the identity order. Accuracy scores are averaged over 10 runs. \label{fig:order_matter}}
    \includegraphics[width=\linewidth]{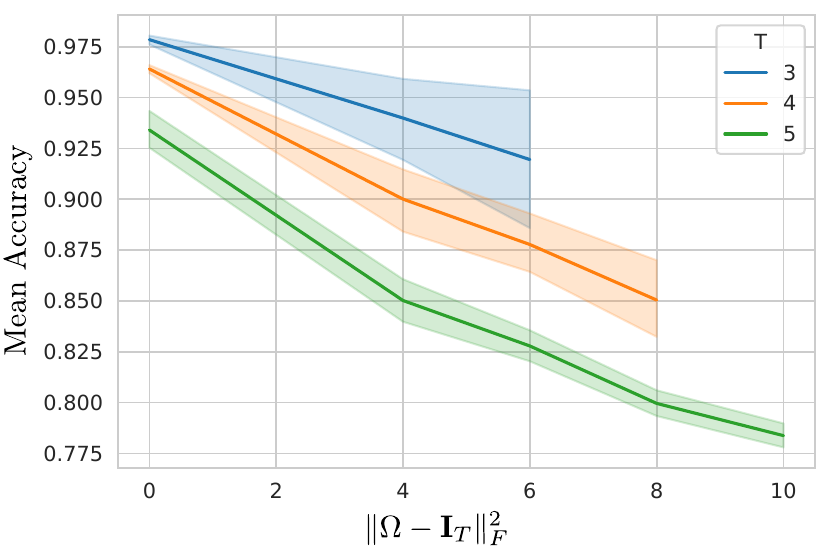}
\end{figure}

\noindent\tac{\textbf{On order importance in toy setting:}} Figure \ref{fig:order_matter} shows the performance of MONET trained with a single, imposed order $\sigma$ on the toy dataset with $T=3,4,5$. For all considered $T$, MONET reaches its best performance when $\sigma$ verifies $\|\mathbf{M}_{\sigma} - \mathbf{I}_{T}\|_{F}^{2} = 0$, which only occurs when the imposed order $\sigma$ is the identity order. Thus, the identity order is the optimal order in the toy experiment. Furthermore, we observe a significant decrease in performance as the imposed order $\sigma$ moves away (\textit{i.e.} as expressed by the frobenius norm of the difference) from the identity order. More precisely, the further the chosen $\sigma$ is from the identity, the lower the accuracy of the network accross the $T$ tasks. Therefore, \textit{order matters} in the toy settings and the best performing orders are the identity order and the orders that are close to it.

\begin{table}
\centering
    \caption{Comparison of different order selection mechanism for $T = 5$ and $M = 120$. Performance are averaged over 10 runs.}
    \vspace{0.1cm}
    \label{tab:ablation_study}
    \resizebox{0.9\linewidth}{!}{
\begin{tabular}{|c|c|c|c|c|}
\hline
Warm up & Dropout & Hard Selection & Mean Accuracy & Finds $\mathbf{I}_{T}$ ? \\ 
\hline
\ding{55} & \ding{55} & \checkmark & 74.2 & 0/10 \\
\ding{55} & \ding{55} & \ding{55} & 89.2 & 8/10 \\
\checkmark & \ding{55} & \ding{55} & 89.3& 9/10\\
\ding{55} & \checkmark & \ding{55}  & 89.8 & 10/10 \\  
\checkmark & \checkmark & \ding{55} & \textbf{90.1} & 10/10 \\  
\hline
\end{tabular}}

\end{table}

\noindent\tac{\textbf{Ablation study on MONET individual components:}} Table \ref{tab:ablation_study} displays the performance of MONET with different order selection mechanism versions. It reports mean accuracy averaged on 10 runs along with the number of times MONET selected the identity order (correct order in this setting) over these runs. As mentioned in the previous paragraph, those two quantities depend from one another because selecting the correct order results in higher performance. 

The worst performing method, that we called Hard Selection is an implementation of \cite{vinyals2015order} order selection mechanism. It uses an exploration phase similar to our warm up and then samples a single order using a distribution in which each order has a probability that is proportional to its performance (measured during exploration). Table \ref{tab:ablation_study} shows that this order selection mechanism perform poorly. In fact, we believe that such sampling strategy is sensitive to performance measurements noise and is therefore likely to result in a sub-optimal order. By smoothly learning its order selection coefficients all along the training phase, MONET gets a better estimation of each order performance and selects better orders.

Refining the order selection mechanism with warm up helps MONET to find better orders by providing an exploration phase to estimate each order performance before making a choice. However it doesn't prevent the order selection from snowballing, i.e to keep allocating weights to the first selected order. It explains that adding warm up to MONET only provides a tiny boost in performance.

\begin{figure*}
\centering
    \input{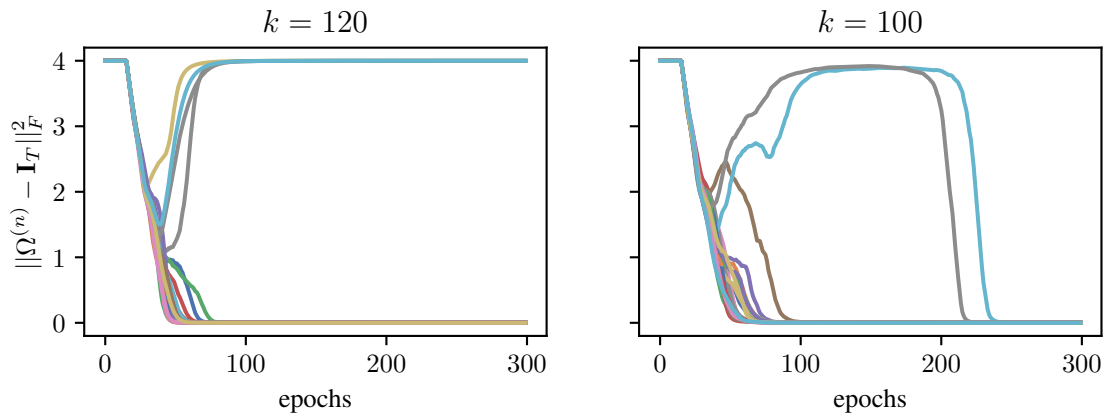}
    \caption{Evolution of the Frobenius distance between MONET soft order and the toy correct order for 20 MONET training with different values of $k$. \tac{Each line represent the soft order trajectory w.r.t the identity matrix for a different training of MONET.  Using the proposed order dropout (k=100, right plot) avoids order snowballing and systematically reach the correct order, as indicated by the Frobenius order falling to 0.}\label{fig:frob2Id}}
\end{figure*}

Dropout on the other hand, short-circuit  the snowball effect by forcing the network to train on randomly selected orders: it allows MONET order selection mechanism to smoothly deviate from any previous choice of order. Indeed, figure \ref{fig:frob2Id}, shows that contrary to its standard version $(k=120, M=120)$, the dropout version of MONET $(k=100, M=120)$ is able to recover the correct order even if it started with a bad guess. Combined warmup and dropout provide MONET with an initial order guess that is likely to be good, and the ability to move away from this guess if needed. Those two ingredients result in better order selection which, in turn, implies better performance.


\begin{figure}
    \centering
\begin{tikzpicture}

\definecolor{color0}{rgb}{0.12156862745098,0.466666666666667,0.705882352941177}
\definecolor{color1}{rgb}{1,0.498039215686275,0.0549019607843137}
\definecolor{color2}{rgb}{0.172549019607843,0.627450980392157,0.172549019607843}
\definecolor{color3}{rgb}{0.83921568627451,0.152941176470588,0.156862745098039}
\definecolor{color4}{rgb}{0,0,0}
\begin{axis}[
legend cell align={left},
legend style={fill opacity=0.8, draw opacity=1, text opacity=1, draw=white!80!black},
tick align=outside,
axis line style = thick,
axis x line=top,
tick pos=left,
x grid style={white!69.0196078431373!black},
x axis line style=color3,
xlabel={\textcolor{color3}{M}},
xmin=-0.5, xmax=5.5,
xtick style={color=black},
xtick={0,1,2,3,4,5},
xticklabels={20,40,60,80,100,120},
y grid style={white!69.0196078431373!black},
ylabel={Accuracy},
ymin=0.835, ymax=0.96,
ytick style={color=black},
yticklabel pos=left,
]

\addplot[only marks,mark=triangle*,mark options={fill=color3, draw=color3, scale=1}] table[x index=0,y index=1,col sep=comma]{%
x  y
0, 0.856159996986389
1, 0.880879992246628
2, 0.908480000495911
3, 0.892960000038147 
4, 0.918479979038239 
5, 0.925440001487732
};
\addlegendentry{no dropout}:

\addplot[only marks,mark=*,mark options={fill=color2, draw=color2, scale=1}] table[x index=0,y index=1,col sep=comma]{%
x  y
0, 0.876559990644455
1, 0.896399992704392
2, 0.919919991493225
3, 0.927279990911484
4, 0.931120002269745
5, 0.925440001487732
};
\addlegendentry{dropout};

\addplot [line width=1.08pt, color3, dashed, forget plot]
table {%
0 0.856159996986389
1 0.880879992246628
2 0.908480000495911
3 0.892960000038147
4 0.918479979038239
5 0.925440001487732
};

\addplot [line width=1.08pt, color3, forget plot]
table {%
0 0.830878009796143
0 0.890725985914469
};
\addplot [line width=1.08pt, color3, forget plot]
table {%
1 0.851599978804588
1 0.908644008785486
};
\addplot [line width=1.08pt, color3, forget plot]
table {%
2 0.883038010150194
2 0.93216400206089
};
\addplot [line width=1.08pt, color3, forget plot]
table {%
3 0.873439991474152
3 0.911603970676661
};
\addplot [line width=1.08pt, color3, forget plot]
table {%
4 0.898075980246067
4 0.93791998103261
};
\addplot [line width=1.08pt, color3, forget plot]
table {%
5 0.903115999251604
5 0.943850026130676
};

\addplot [line width=1.08pt, color2, dashed, forget plot]
table {%
0 0.876559990644455
1 0.896399992704392
2 0.919919991493225
3 0.927279990911484
4 0.931120002269745
5 0.925440001487732
};
\addplot [line width=1.08pt, color2, forget plot]
table {%
0 0.871680003255606
0 0.880722007006407
};
\addplot [line width=1.08pt, color2, forget plot]
table {%
1 0.89095798432827
1 0.901919973045587
};
\addplot [line width=1.08pt, color2, forget plot]
table {%
2 0.913917991220951
2 0.926081987917423
};
\addplot [line width=1.08pt, color2, forget plot]
table {%
3 0.921277979016304
3 0.932882013469934
};
\addplot [line width=1.08pt, color2, forget plot]
table {%
4 0.916393992006779
4 0.943522008955479
};
\addplot [line width=1.08pt, color2, forget plot]
table {%
5 0.905833979547024
5 0.94320602118969
};
\addplot [semithick, color4, forget plot]
table {%
-0.5 0.948
5.5 0.948
};
\addlegendentry{oracle};
\addplot [semithick, dashed, color4, forget plot]
table {%
-0.5 0.845
5.5 0.845
};
\addlegendentry{baseline};
\end{axis}

\begin{axis}[
axis x line=top,
x axis line style=color2,
tick align=outside,
x grid style={white!69.0196078431373!black},
axis line style = thick,
ytick pos=left,
xtick={0,1,2,3,4,5},
xticklabels={20,40,60,80,100,120},
y grid style={white!69.0196078431373!black},
ylabel={Accuracy},
ymin=0.835, ymax=0.96,
ytick style={},
yticklabel pos=left,
xlabel={\textcolor{color2}{k}},
xmin=-0.5, xmax=5.5,
xtick style={color=black}
]
\end{axis}

\end{tikzpicture}
    \caption{MONET performances comparaison for toy dataset with $T=5$ between: (in red) dropout less version (k=M) for different values of $M$  and (in green) dropout version with M=T! and different values of $k$. Dashed and full black lines are respectively the random order selection baseline and the oracle mean performances. Means and standard deviations are computed on 10 runs.}
    \label{fig:k_beats_M}
\end{figure}
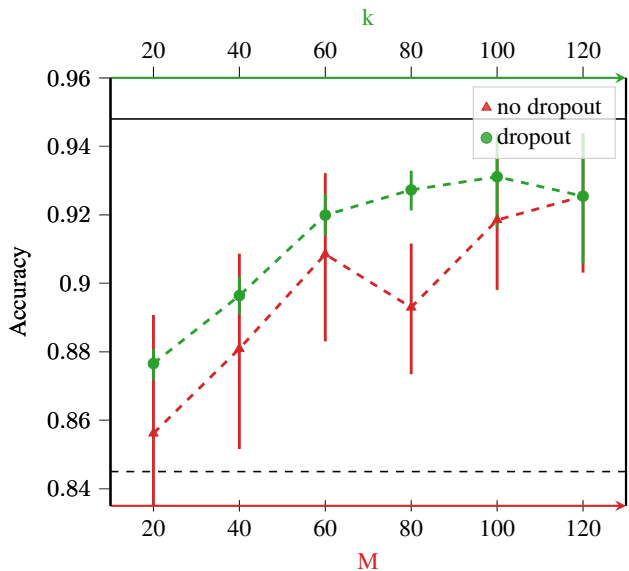

\noindent\tac{\textbf{Hyperparameter tuning:} Figure \ref{fig:k_beats_M}, compares the performance of MONET for $T=5$ tasks with different settings for $k$ and $M$. First, it shows that the performance of MONET without order dropout (red plot) are increasing with the number of randomly sampled order $M$. This is fairly logical. Indeed, the higher $M$, the more likely it is to get a good order in the set of randomly selected orders, the better the performance are. Moreover, it confirms that MONET with $M=120$ and order dropout (green plot) achieve higher accuracy and get closer to the oracle with enforced correct order $\mathbf{I}_{T}$.}



\begin{table}
	\caption{Multi-task baselines comparaison on toy dataset with $T=5$ tasks. $^\dagger$: oracle predictor with correct order. \label{tab:toybaseline}}
	\vspace{0.1cm}
	\resizebox{\linewidth}{!}{
\begin{tabular}{|c|c|c|c|c|c|c|c|c|c|c|c|}

\hline
\textbf{Accuracy} & Task 1 & Task 2 & Task 3 & Task 4 & Task 5 & Mean \\
\hline
VMNC  & $98.7 $ & $97.6$ & $92.4$ & $65.2$ & $54.0$ & $81.5$ \\
VMNS & $ 98.7 $ & $ 97.5$ & $ 72.8$ & $ 51.5$ & $52.9$ & $74.9$\\
MRNN & $93.7$ & $93.6$ & $80.1$ & $60.5$ & $51.3$ & $75.9$ \\
MONET & $99.4$ & $98.3$ & $95.8$ & $90.0$ & $68.2$ & $90.1$ \\
\hline
\rowcolor{lightgray} Oracle$^\dagger$ & $99.6$ & $98.6$ & $96.1$ & $92.1$ & $80.6$  & $93.4$ \\
\hline

\end{tabular}}
\end{table}

\noindent\tac{\textbf{Comparison with baseline methods and discussion:}} Lastly, Table \ref{tab:toybaseline} shows relative performances of MONET w.r.t multi-task baselines.
MONET displays significantly better performances than all other considered methods, significantly narrowing the gap with the oracle performance.

Eventually, we demonstrated in a controlled benchmark where an optimal task chaining order is known that (a) MONET was able to consistently retrieve said order, and (b) that thanks to its joint order selection mechanism and task-specific recurrent cell sharing architecture, backed by the proposed order dropout strategy, MONET was able to consistently outperform other multi-task baselines, getting closer to an oracle predictor using the optimal order. We now consider real-world applications with potentially more complex inter-task dependencies and less clear ordering patterns.

\subsection{Applications to face attribute detection}
Table \ref{tab:celeba_results} draws a comparison between different multi-task methods for attribute detection on CelebA. On the one hand, there is no clear winner between the two VMN versions: for instance, VMNS performs better on the \textit{gender} and \textit{accessories} subsets while VMNC performs better on \textit{haircut} and \textit{misc.}. Those performance discrepancies may result in practical difficulties to find an all-around, well performing architecture, as echoed in \cite{misra2016cross}. Furthermore, MRNNs gets consistently outperformed by at least one of the VMN methods. In fact, we believe that MRNN recurrent cell sharing across tasks leads to early conflicts between task-associated gradients and prevents it from taking full advantage of its theoretically better inter-task relationship modelling. Additionally, random task order sampling may prevent MRNN from properly learning and hurt its predictive performance.  MONET, on the other hand, shows consistently better performances than both VMN as well as MRNN on every subset, due to both its task-wise modular weight sharing strategy and its order selection mechanism that, in turn, allows to correctly model inter-task dependencies.

\begin{table}[ht]
	\caption{Comparaison of MONET with multi-task baselines on several attributes subsets of CelebA. \label{tab:celeba_results}}
	\vspace{0.1cm}
	\centering
\resizebox{1.03\linewidth}{!}{
\begin{tabular}{ |c|c|c|c|c|c|c| } 
 \hline
 \footnotesize Gender & \footnotesize H. Makeup & \footnotesize Male & \footnotesize Mustache & \footnotesize No Beard & \footnotesize W Lipstick & Avg. \\
 \hline
 VMNC & $88.0 $ & $95.8 $ & $96.6 $ & $95.4 $ & $90.2 $ & $93.2 $ \\
 VMNS & $90.2 $ & $97.0 $ & $96.7 $ & $95.5 $ & $93.9 $ & $94.6 $ \\
 MRNN & $89.9 $ & $96.9 $ & $96.7 $ & $95.6 $ & $93.6 $ & $94.5 $ \\
 MONET & $90.5 $ & $97.5 $ & $96.8 $ & $95.8 $ & $93.9 $ & $\mathbf{94.9}$ \\
 \hline
 \hline
\footnotesize Accessories & \footnotesize Eyeglasses & \footnotesize W Earrings &\footnotesize W Hat &\footnotesize W Necklace &\footnotesize W Necktie & Avg. \\
 \hline
 VMNC & $99.4 $ & $88.5 $ & $98.4 $ & $86.8 $ & $95.8 $ & $93.8 $\\
 VMNS & $99.5 $ & $89.5 $ & $98.6 $ & $87.1 $ & $96.7 $ & $\mathbf{94.3 }$ \\
 MRNN & $99.5 $ & $89.6 $ & $98.7 $ & $87.1 $ & $96.6 $ & $\mathbf{94.3 }$ \\
 MONET & $99.2 $ & $89.9 $ & $98.5 $ & $87.3 $ & $96.8 $ & $\mathbf{94.3 }$ \\
 \hline
 \hline
\footnotesize Haircut &\footnotesize Bald &\footnotesize Black Hair &\footnotesize Blond Hair &\footnotesize Brown Hair &\footnotesize Gray Hair & Avg. \\
 \hline
 VMNC & $98.6 $ & $88.3 $ & $95.3 $ & $88.0 $ & $98.0 $ & $93.6 $\\
 VMNS & $98.6 $ & $86.2 $ & $95.2 $ & $86.9 $ & $97.9 $ & $93.0 $ \\
 MRNN & $98.6 $ & $86.7 $ & $95.2 $ & $86.2 $ & $97.8 $ & $92.9 $ \\
 MONET & $98.7 $ & $88.3 $ & $95.4 $ & $88.0 $ & $98.0 $ & $\mathbf{93.7 }$ \\
 \hline
 \hline
\footnotesize Beauty &\footnotesize A Eyebrows &\footnotesize Attractive &\footnotesize H ChBones &\footnotesize R Cheeks &\footnotesize Oval Face & Avg. \\
 \hline
 VMNC & $82.8 $ & $81.1 $ & $86.8 $ & $94.2 $ & $73.7 $ & $83.7 $ \\
 VMNS & $82.5 $ & $81.0 $ & $86.5 $ & $94.4 $ & $74.0 $ & $83.7 $ \\
 MRNN & $82.8 $ & $80.6 $ & $86.3 $ & $94.6 $ & $74.1 $ & $83.7 $ \\
 MONET & $82.8 $ & $81.5 $ & $86.9 $ & $94.7 $ & $74.4 $ & $\mathbf{84.1 }$ \\
 \hline
 \hline
\footnotesize Misc. &\footnotesize  5 Shadow &\footnotesize P Nose &\footnotesize M S Open &\footnotesize Oval Face &\footnotesize Young & Avg. \\
 \hline
 VMNC & $93.6 $ & $76.6 $ & $93.5 $ & $74.2 $ & $86.4 $ & $84.9 $ \\
 VMNS & $93.6 $ & $76.7 $ & $84.8 $ & $74.1 $ & $86.8 $ & $83.2 $ \\
 MRNN & $93.6 $ & $76.1 $ & $93.2 $ & $73.6 $ & $85.6 $ & $84.4 $ \\
 MONET & $94.2 $ & $76.7 $ & $93.6 $ & $74.3 $ & $87.0 $ & $\mathbf{85.2 }$ \\
 \hline
\end{tabular}
}

\end{table}

\begin{figure}
    \centering
    \includegraphics[width=\linewidth]{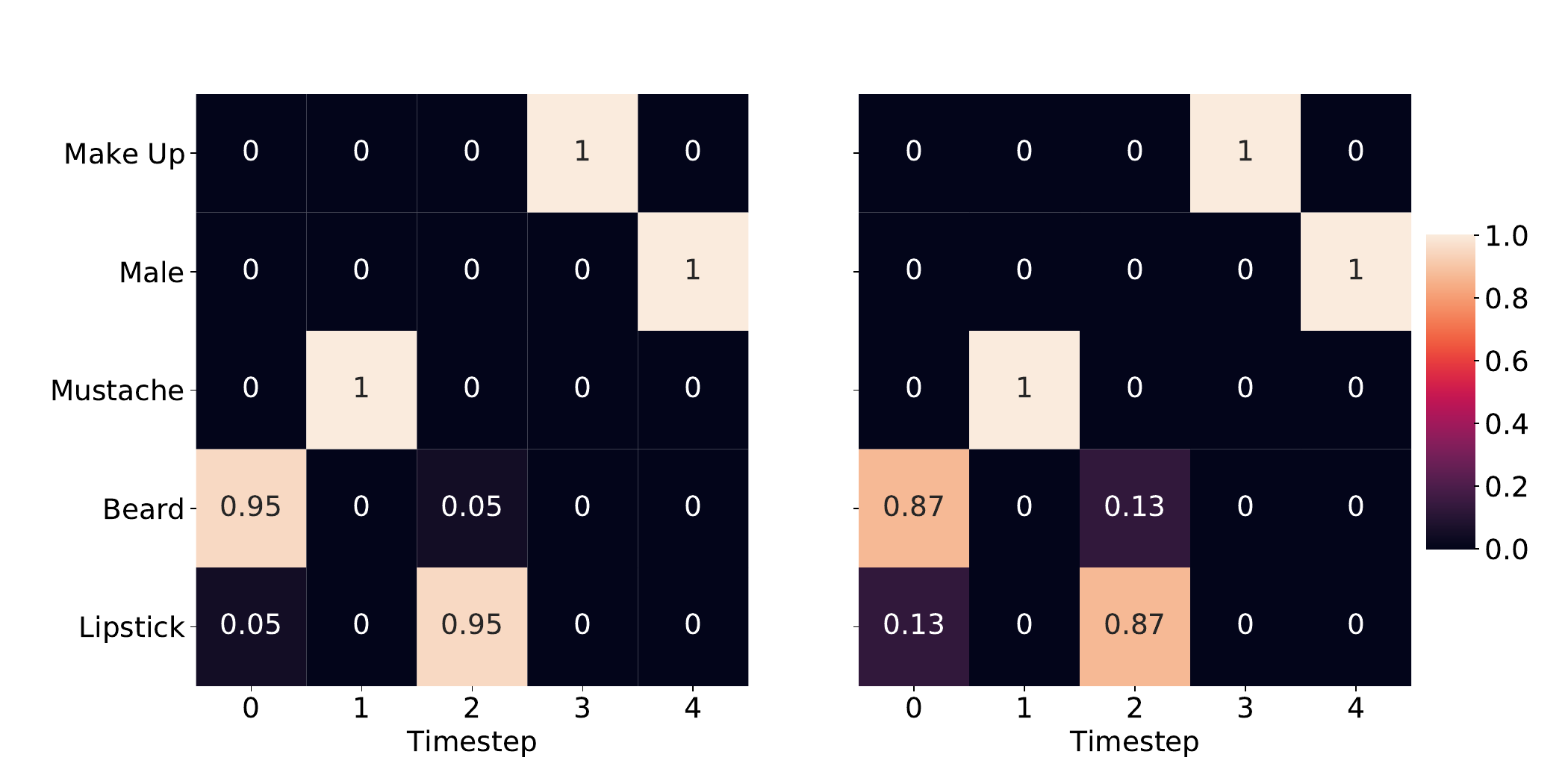}
    \caption{Two soft-order matrices extracted at the end of two different MONET training on CelebA \textit{gender} subset.     \label{fig:celeba_matrices}}
\end{figure}

Figure \ref{fig:celeba_matrices} depicts two soft-order matrices extracted at the end of MONET training on the \textit{gender} subset. First, those two matrices are very similar, showing that MONET order selection mechanism is relatively stable across several networks and order selector initializations. \tac{Second, it seems that MONET learns to process easy tasks in priority and uses the result of those easy tasks to condition the prediction on harder ones. For example, it typically learns to predict \textit{beard} (which is more visible and therefore easier to predict) before predicting \textit{mustache} and \textit{lipstick} (which has a very characteristic color) before \textit{heavy makeup} (which exhibit more variability and is fairly subjective). From an intuitive point of view, such learned order is fairly reasonable. Indeed, predicting easy tasks earlier  reduces the chances of propagating prediction mistakes along the processing chain and in turn improves performance.}

\subsection{Applications to action unit detection}

\begin{table*}[ht]
	\caption{Comparaison of MONET with state-of-the-art deep learning based AU detection methods on BP4D \label{tab:bp4d_results}}
	\resizebox{1.03\linewidth}{!}{
\begin{tabular}{|c|c|c|c|c|c|c|c|c|c|c|c|c|c|c|}

\hline
\footnotesize {F1 Score-AU} &1 & 2 & 4 & 6 & 7 & 10 & 12 & 14 & 15 & 17 & 23 & 24 & {Avg.} \\
\hline
\footnotesize DRML \cite{zhao2016deep} & 36.4 & 41.8 & 43.0 & 55.0 & 67.0 & 66.3 & 65.8 & 54.1 & 33.2 & 48.0 & 31.7 & 30.0 & 48.3 \\
\footnotesize EAC-NET \cite{li2018eac} & 39.0 & 35.2 & 48.6 & 76.1 & 72.9 & 81.9 & 86.2 & 58.8 & 37.5 & 59.1 & 35.9 & 35.8 & 55.9 \\
\footnotesize DSIN \cite{corneanu2018deep} & 51.7 & 40.4 & 56.6 & 76.1 & 73.5 & 79.9 & 85.4 & 62.7 & 37.3 & 62.9 & 38.8 & 41.6 & 58.9 \\
\footnotesize JAANet \cite{shao2018deep} & 47.2 & 44.0 & 54.9 & 77.5 & 74.6 & 84.0 & 86.9 & 61.9 & 43.6 & 60.3 & 42.7 & 41.9 & 60.0 \\
\footnotesize LP-Net \cite{Niu_2019_CVPR} & 43.4 & 38.0 & 54.2 & 77.1 & 76.7 & 83.8 & 87.2 & 63.3 & 45.3 & 60.5 & 48.1 & 54.2 & 61.0 \\
\footnotesize CMS \cite{sankaran2019representation} & 49.1 & 44.1 & 50.3 & 79.2 & 74.7 & 80.9 & 88.3 & 63.9 & 44.4 & 60.3 & 41.4 & 51.2 & 60.6 \\
\footnotesize ARL \cite{shao2019facial} & 45.8 & 39.8 & 55.1 & 75.7 & 77.2 & 82.3 & 86.6 & 58.8 & 47.6 & 62.1 & 47.7& 55.4 & 61.1 \\
\footnotesize SRERL \cite{li2019semantic} & 46.9 & 45.3 & 55.6 & 77.1 & 78.4 & 83.5 & 87.6 & 63.9 & 52.2 & {63.9} & 47.1 & 53.3 & 62.1 \\
\footnotesize JÂANET \cite{shao2020jaa} & 53.8 & 47.8 & 58.2 & 78.5 & 75.8 & 82.7 & 88.2 & 63.7 & 43.3 & 61.8 & 45.6 & 49.9 & 62.4 \\
\footnotesize HMP-PS \cite{song2021hybrid} & 53.1 & 46.1 & 56.0 & 76.5 & 76.9 & 82.1 & 86.4 & 64.8 & 51.5 & 63.0 & 49.9 & 54.5 & 63.4 \\
\footnotesize SEV-Net \cite{yang2021exploiting} &  58.2 & 50.4 & 58.3 & 81.9 & 73.9 & 87.8 & 87.5 & 61.6 & 52.6 & 62.2 & 44.6 & 47.6 & 63.9 \\
\footnotesize PT-MT-ATsup-CC-E \cite{jacob2021facial} & 51.7 & 49.3 & 61.0 & 77.8 & 79.5 & 82.9 & 86.3 & 67.6 & 51.9 & 63.0 & 43.7 & 56.3 & 64.2 \\
\hline
\tac{VMNS} & 51.7 & 46.6 & 57.8 & 77.7 & 74.2 & 81.1 & 88.3 & 59.3 & 45.7 & 60.8 & 45.0 & 49.5 & 61.5 \\
\tac{VMNC} & \tac{48.7} & \tac{45.2} & \tac{56.8} & \tac{77.9} & \tac{77.8} & \tac{83.2} & \tac{87.9} & \tac{62.9} & \tac{51.1} & \tac{59.1} & \tac{47.4} & \tac{52.7} & \tac{62.6} \\
\tac{MRNN} & \tac{47.1} & \tac{44.7} & \tac{59.1} & \tac{77.5} & \tac{78.3} & \tac{84.1} & \tac{85.3} & \tac{63.9} & \tac{41.6} & \tac{62.8} & \tac{43.3} & \tac{51.5} & \tac{61.6} \\
MONET (ours) & 54.5 & 45.0 & 61.5 & 75.9 & 78.0 & 84.5 & 87.6 & 65.1 & 54.8 & 60.5 & 53.0 & 53.2 & \textbf{64.5} \\

\hline
\end{tabular}
}
\end{table*}

\textbf{AU detection on BP4D:} Table \ref{tab:bp4d_results} draws a comparison between MONET and other state of the art deep approaches on the detection of 12 action units on BP4D database. Thanks to its order selection mechanism, MONET outperforms all the methods that explicitly models AU label dependencies such as DSIN or HMP-PS. More interestingly, it outperforms methods that use external information such as landmarks (EAC-NET, JAANET, PT-MT-ATsup-CC-E) or textual description of Action Units (SEV-NET). Finally, MONET performs better than PT-MT-ATsup-CC-E which leverages transformers in its architecture. Thus, despite the fact that there is considerable room for improvement, MONET outperforms existing approaches due to its ability to jointly model task order and prediction.

\begin{figure*}
\centering
    \includegraphics[width=\linewidth]{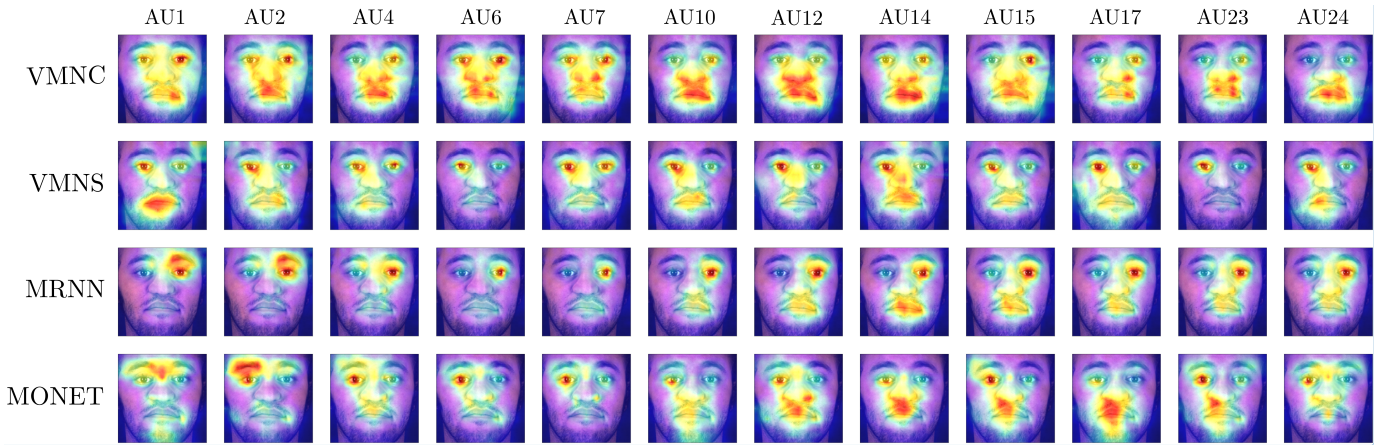}
    \vspace{-0.5cm}
    \caption{\tac{AU-wise attribution maps for multi-task baselines as well as MONET. While other architectures may stuggle to correctly locate each AU, MONET qualitatively retrieve the correct area for each AU due to parameter sharing and joint task distribution modelling.}}
    \label{figure:att_map}
\end{figure*}

\tac{Figure \ref{figure:att_map} display AU-wise attribution maps for the three multi-task baselines along with MONET. Those maps are computed using Grad-Cam-like \cite{selvaraju2017grad} techniques which consists in taking the gradients of each action unit prediction with respect to the input image or intermediate feature maps. VMNC and MRNN predicts action units using an architecture that is fully shared across tasks. Therefore, there is no space for task specialization which result in similar heatmaps across tasks. For example, MRNN heatmaps from AU14 to AU24 are all very close. On the other hand, VMNS which uses a specific regressor by task manage to specialize its attribution for each tasks (heatmaps by tasks are different). However it misses important localizations such as eyebrows zone for AU1, AU2 while MRNN catches them. In fact, we believe that the joint task distribution modelling in MRNN helps guide the network attention. MONET gets the best of the two worlds: On one side, its task-wise module enable task-wise attribution specialization, on the other side its joint task distribution modelling helps guide the attribution toward specific zones such as eyebrows for AU1-2.}

\tac{Finally Table \ref{tab:bp4d_results} shows a comparison between MONET and the other multi-task baselines. First, we observe a large gap in performance between VMNC and VMNS. It demonstrates that weight sharing pattern choices critically affect performance. More importantly it shows that choosing the best sharing pattern for a task is a difficult problem. Second, MRNN performance lies lower than VMNC. We believe that this is due to the larger number of tasks, that makes it less likely to find a suitable order to efficiently learn the Action Unit dependencies. Lastly, thanks to its task-wise modular sharing pattern and order selection mechanism, MONET efficiently models the joint AU distribution and achieves better performance than the other three methods.}

\textbf{AU detection on DISFA:} Table \ref{tab:disfa_results} compares the performance of  MONET with other state of the art approaches on DISFA. \tac{Similar to BP4D, MONET displays better performance than classical multi-task methods and all other existing approaches}. Therefore, MONET consistently outperform state of the art performance for facial action unit recognition.

\begin{table*}[ht]
	\caption{Comparaison of MONET with state-of-the-art deep learning based AU detection methods on DISFA \label{tab:disfa_results}}
	\vspace{0.1cm}
	\centering
\resizebox{0.82\linewidth}{!}{
\begin{tabular}{|c|c|c|c|c|c|c|c|c|c|c|c|}

\hline
\footnotesize \textbf{F1 Score-AU} &1 &2 & 4 & 6 & 9 & 12 & 25 & 26 & \textbf{Avg.} \\
\hline
\footnotesize DRML \cite{zhao2016deep} & 17.3 & 17.7 & 37.4 & 29.0 & 10.7 & 37.7 & 38.5 & 20.1 & 26.7 \\
\footnotesize EAC-NET \cite{li2018eac} & 41.5 & 26.4 & 66.4 & 50.7 & 8.5 & 89.3 & 88.9 & 15.6 & 48.5 \\
\footnotesize DSIN \cite{corneanu2018deep} & 42.4 & 39.0 & 68.4 & 28.6 & 46.8 & 70.8 & 90.4 & 42.2 & 53.6 \\
\footnotesize SRERL \cite{li2019semantic} & 45.7 & 47.8 & 59.6 & 47.1 & 45.6 & 73.5 & 84.3 & 43.6 & 55.9 \\
\footnotesize JAANet \cite{shao2018deep} & 43.7 & 46.2 & 56.0 & 41.4 & 44.7 & 69.6 & 88.3 & 58.4 & 56.0 \\
\footnotesize LP-Net \cite{Niu_2019_CVPR} & 29.9 & 24.7 & 72.7 & 46.8 & 49.6 & 72.9 & 93.8 & 65.0 & 56.9 \\
\footnotesize CMS \cite{sankaran2019representation} & 40.2 & 44.3 & 53.2 & 57.1 & 50.3 & 73.5 & 81.1 & 59.7 & 57.4 \\
\footnotesize ARL \cite{shao2019facial} & 43.9 & 42.1 & 63.6 & 41.8 & 40.0 & 76.2 & 95.2 & 66.8 & 58.7 \\
\footnotesize SEV-Net \cite{yang2021exploiting} & 55.3 & 53.1 & 61.5 & 53.6 & 38.2 & 71.6 & 95.7 & 41.5 & 58.8 \\
\footnotesize HMP-PS \cite{song2021hybrid} & 38.0 & 45.9 & 65.2 & 50.9 & 50.8 & 76.0 & 93.3 & 67.6 & 61.0 \\
\footnotesize PT-MT-ATsup-CC-E \cite{jacob2021facial} & 46.1 & 48.6 & 72.8 & 56.7 & 50.0 & 72.1 & 90.8 & 55.4 & 61.5 \\
\footnotesize JÂANET \cite{shao2020jaa} & 62.4 & 60.7 & 67.1 & 41.1 & 45.1 & 73.5 & 90.9 & 67.4 & 63.5 \\
\hline
\tac{VMNS} & \tac{53.4} & \tac{51.3} & \tac{64.8} & \tac{45.5} & \tac{36.0} & \tac{70.1} & \tac{89.8} & \tac{62.4} & \tac{59.2} \\
\tac{VMNC} & \tac{56.8} & \tac{59.0} & \tac{64.4} & \tac{51.4} & \tac{43.7} & \tac{75.1} & \tac{92.5} & \tac{62.8} & \tac{63.2} \\
\tac{MRNN} & \tac{47.4} & \tac{49.7} & \tac{61.8} & \tac{46.7} & \tac{38.8} & \tac{71.0} & \tac{91.9} & \tac{60.9} & \tac{58.5} \\ 
MONET (ours) & 55.8 & 60.4  & 68.1 & 49.8 & 48.0 & 73.7 & 92.3 & 63.1 & $\mathbf{63.9}$ \\
\hline
\end{tabular}
}

\end{table*}

\begin{figure}
    \centering
    \includegraphics[width=\linewidth]{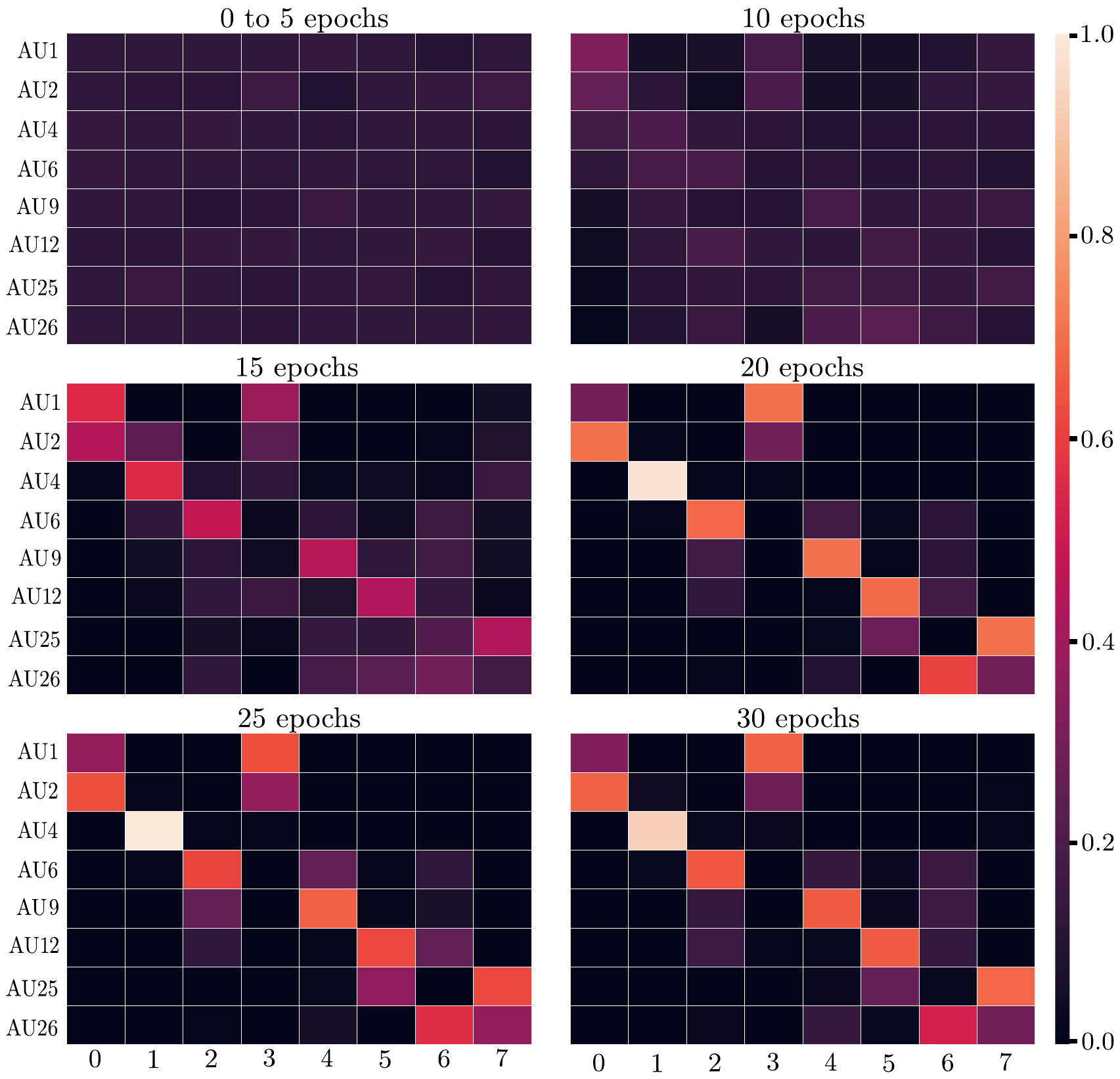}
    \caption{Evolution of MONET soft order during 30 epochs of training on DISFA. For action unit $j$, the coefficient associated with timestep $i$ can be interpreted as the probability that action unit $j$ is processed at timestep $i$. \label{fig:disfa_matrices}}
\end{figure}

Figure \ref{fig:disfa_matrices} shows MONET soft order evolution when training on DISFA. During the warm up phase (first 5 epochs), the soft order is the average of $512$ randomly sampled permutation matrices and is therefore close to uniform (all probabilities are close to $\frac{1}{T}$). \tac{At the end of the warm up, MONET starts learning the order selection coefficients. In particular, we observe that MONET learns to start by predicting AU1, 2, 4, 6, 9 which correspond to the upper part of the face (eyes and brows related action units) and then processes 12, 25 and 26 which are located in the lower part of the face (mostly the mouth). In fact we believe that tasks that use the same part of the image and more generally the same piece of information benefit from being processed close to each other.}

This coarse intuition seems to be fairly verified at a more detailed scale as we also observe smaller blocks of consecutive tasks. For instance, AU25-26 basically focuses on mouth and jaw, and AU4-6 both have their regions of interest located around the eyes. As far as brow movements are concerned, it is worth noticing that AU4 is processed before AU1. An explanation could be that AU4 is easier to predict (mainly because it is less local as it often comes with nose and corner of the eye wrinkles). Therefore using AU4 prediction and learned dependencies to help predict AU1 is more beneficial than doing the opposite.

\section{Conclusion and Discussion}
\label{sec:conclusion}
\tac{In this paper, we tackled the problem of AU detection, which is intrinsically a multi-task problem with strong inter-task dependencies.}

To efficiently model these relationships, we introduced MONET, a multi-order network for joint task order and prediction modelling in deep multi-task learning. MONET leverages a differentiable order selection mechanism based on soft order modelling inside Birkhoff's polytope, as well as task-wise recurrent cell sharing for concurrent multi-order prediction learning. Furthermore, we propose warmup and order dropout strategies that enhance order selection by preventing order overfitting.

Experimentally, we first demonstrated that MONET was able to converge toward the correct order in a controlled scenario. Second, we showed that demonstrate that MONET display competitive performance on a wide variety of task relationship settings: applied to facial attribute detection (CelebA), MONET performs at least as well as the best multi-task baseline on each 5 attributes subset. Based on this empirical evidence, we argue that MONET is an all-around better multi-task method and could therefore constitute a valid architectural choice for many multi-task learning problems. 



\tac{Then, we demonstrated that, thanks to its order selection mechanism and task-wise modular architecture, MONET efficiently models the strong inter-task dependencies between AUs and consequently outperform multi-task baselines aswell as state-of-the art performance on both DISFA and BP4D datasets.}

Finally, the proposed work still suffers from certain limitations. The most important is the scalability to a large number of tasks. Indeed, the number of permutations for soft order modelling increases as the factorial of the number of tasks, and the number of recurrent cells in the architecture grows linearly with the number of tasks. \tac{Furthermore, strong correlations between AUs could be used to partition the set of AUs into several blocks with strong intra-block correlation and low inter-block correlation. Then we could use a recurrent cell by blocks of task and learn the best block order with MONET, which would also address the problem of the numbers of tasks. Furthermore, an interesting direction will consist in extending MONET to other families of tasks (eg: categorical classification, regression) to design efficient multi-task affective computing methods, including face recognition, AU intensity and arousal/valence estimation, as well as facial landmarks and head pose estimation.}
\section{Acknowledgements}
This work was granted access to the HPC resources of IDRIS under the allocation 2021-AD011013183 made by GENCI

\ifCLASSOPTIONcaptionsoff
  \newpage
\fi



%
\bibliographystyle{IEEEtran}
\bibliography{main}

%

\begin{IEEEbiography}[{\includegraphics[width=1in,height=1.25in,clip,keepaspectratio]{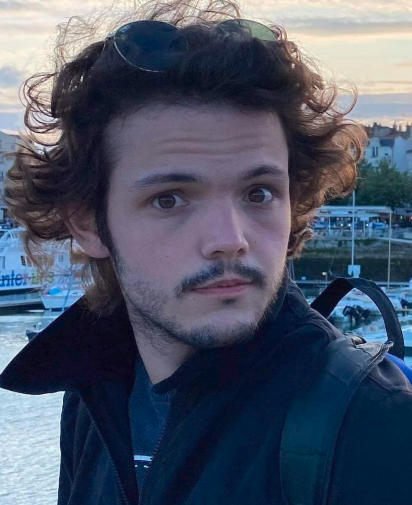}}]{Gauthier TALLEC}
is pursuing a PhD with the Institute of Intelligent Systems and Robotics (ISIR) at Sorbonne University in Paris. He received both the engineering degree from Telecom ParisTech and the M.S.degree in computer science from ENS Paris-Saclay in 2019. His work concerns deep learning and more precisely multi-task deep learning for computer vision. He is particularly interested in affective computing and face analysis applications.
\end{IEEEbiography}

\begin{IEEEbiography}[{\includegraphics[width=1in,height=1.25in,clip,keepaspectratio]{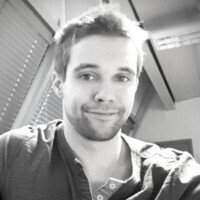}}]{Arnaud DAPOGNY} is a computer vision researcher at Datakalab in Paris. He obtained the Engineering degree from the Sup\'elec engineering School in 2011 and the Masters degree from Sorbonne University, Paris, in 2013 with high honors. He also obtained his PhD at Institute for Intelligent Systems and Robotics (ISIR) in 2016 and worked as a post-doctoral fellow at LIP6. His works concern deep learning for computer vision and its application to automatic facial behavior as well as gesture analysis.
\end{IEEEbiography}


\begin{IEEEbiography}[{\includegraphics[width=1in,height=1.25in,clip,keepaspectratio]{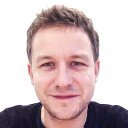}}]{Kevin BAILLY}
is associate professor with the Institute of Intelligent Systems and Robotics (ISIR) at Sorbonne University and Head of Research of Datakalab. He received the PhD degree in computer science from the Pierre et Marie Curie University in 2010 and was a postdoctoral researcher at Telecom Paris from 2010 to 2011. His research interests are in machine learning and computer vision applied to face processing and behavior analysis.
\end{IEEEbiography}




\end{document}